\pdfoutput=1
\RequirePackage[hyphens]{url}
\documentclass[11pt]{article}

\usepackage[preprint]{acl}
\usepackage{lipsum}

\usepackage{import}
\usepackage{multirow}
\usepackage{placeins}
\usepackage{times}
\usepackage{latexsym}
\usepackage{pbox}
\usepackage[T1]{fontenc}

\usepackage[utf8]{inputenc}

\usepackage{microtype}

\usepackage{inconsolata}

\usepackage{colortbl}
\usepackage{graphicx}
\usepackage{xspace}
\usepackage{tabularx}
\usepackage{tabulary}
\usepackage{amsmath}
\usepackage{booktabs}
\usepackage{listings}
\usepackage[most]{tcolorbox}
\usepackage{colortbl}
\usepackage{cuted}
\usepackage{xcolor}
\usepackage{kotex}
\usepackage{pifont}
\usepackage{dirtytalk}
\usepackage{algorithm}
\usepackage{algpseudocode}
\usepackage[utf8]{inputenc}
\usepackage[T1]{fontenc}
\usepackage[perpage]{footmisc}
\usepackage{hyperref}
\usepackage{enumitem}

\usepackage{ulem}
\usepackage{makecell}

\usepackage{amsmath, amssymb}
\usepackage[most]{tcolorbox}
\usepackage{siunitx}
\usepackage{graphicx, xcolor, colortbl, booktabs}
\usepackage{array, tabularx}
\usepackage{caption}
\usepackage{subcaption}
\usepackage{float}

\definecolor{green}{RGB}{144, 238, 144}
\definecolor{red}{RGB}{255, 102, 102}

\newcommand{\up}[1]{\textcolor{green}{\textbf{#1} ▲}}
\newcommand{\down}[1]{\textcolor{red}{\textbf{#1} ▼}}

\newtcolorbox{instructionsbox}[1][]{
  colframe=cyan!75!black,    
  colback=green!5!white,     
  coltitle=black,            
  title=#1,                  
  rounded corners,           
  boxrule=0.5mm,             
  boxsep=5pt,                
  toptitle=1mm,              
  bottomtitle=1mm,           
  left=0pt,                 
  right=0pt,                
  top=0pt,                   
  bottom=0pt,                
  fonttitle=\bfseries        
}

\newcommand{\traindataset}{\textsc{Euler-Instruct}\xspace}

\title{Linguistic Generalizability of Test-Time Scaling in Mathematical Reasoning}

\author{
Guijin Son{\textsuperscript{1,2}} \quad Jiwoo Hong{\textsuperscript{3}} \quad Hyunwoo Ko{\textsuperscript{2}} \quad James Thorne{\textsuperscript{3}}  
 \\ \\
Yonsei University{\textsuperscript{1}} \quad OneLineAI{\textsuperscript{2}} \quad KAIST AI{\textsuperscript{3}}
\\
\texttt{spthsrbwls123@yonsei.ac.kr}\\ 
}

\begin{document}
\maketitle

\begin{abstract}
  
Scaling pre-training compute has proven effective for achieving multilinguality, but does the same hold for test-time scaling? In this work, we introduce \textbf{MCLM}, a multilingual math benchmark featuring competition-level problems in 55 languages. We test three test-time scaling methods—Outcome Reward Modeling (ORM), Process Reward Modeling (PRM), and Budget Forcing (BF)—on both Qwen2.5-1.5B Math and \textbf{MR1-1.5B}, a multilingual LLM we trained for extended reasoning. Our experiments show that using Qwen2.5-1.5B Math with ORM achieves a score of 35.8 on MCLM, while BF on MR1-1.5B attains 35.2. Although “thinking LLMs” have recently garnered significant attention, we find that their performance is comparable to traditional scaling methods like best-of-N once constrained to similar levels of inference FLOPs. Moreover, while BF yields a 20-point improvement on English AIME, it provides only a 1.94-point average gain across other languages—a pattern consistent across the other test-time scaling methods we studied—highlighting that test-time scaling may not generalize as effectively to multilingual tasks. To foster further research, we release MCLM, MR1-1.5B, and evaluation results.~\footnote{\url{https://github.com/gauss5930/MCLM}}

\end{abstract}

\section{Introduction}

Large Language Models (LLMs) have achieved impressive gains across a wide range of tasks by scaling compute during pre-training~\citep{thoppilan2022lamda, smith2022using}. Contrary to early concerns about a so-called “curse of multilinguality,”~\citep{conneau-etal-2020-unsupervised, pfeiffer-etal-2022-lifting} which suggested that training in diverse languages would degrade overall performance, sufficiently large decoder-only architectures have demonstrated strong multilingual capabilities~\citep{dubey2024llama3, aryabumi2024aya23}. Yet as further scaling becomes increasingly difficult—due to data scarcity~\citep{longpre2024consent}, diminishing returns, or prohibitive costs~\citep{achiam2023gpt}—researchers have begun exploring test-time scaling methods that expand a model’s reasoning or generation capacity at test time. An intriguing question arises: \emph{Does test-time scaling confer the same cross-lingual benefits we see at train-time scaling during pre-training?}

\begin{figure}[t!]
\centering
\includegraphics[width=\columnwidth]{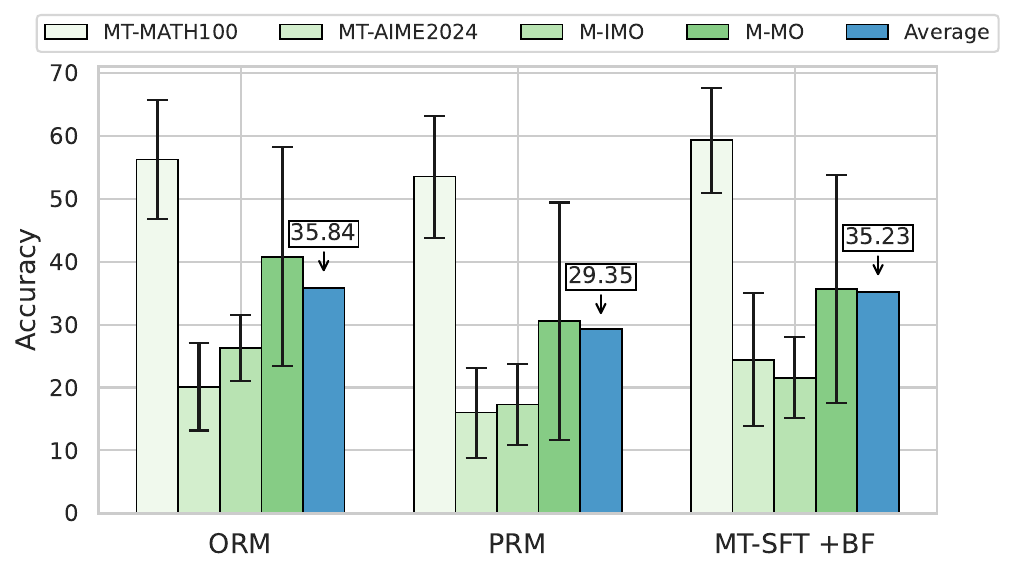}
\vspace{-0.3in}
\caption{\footnotesize \textbf{Performance of Qwen2.5-1.5B-Math with different test-time scaling strategies.}——Once configured to use comparable inference FLOPs, all three methods (Outcome Reward Modeling, Process Reward Modeling, and Budget Forcing) achieve similar performance.}
\vspace{-0.15in}
\label{fig_title}
\end{figure}

Early studies demonstrated that chain-of-thought prompting~\citep{wei2022chain} and scratchpads~\citep{nye2021show} can significantly boost model performance—particularly in mathematics~\citep{lewkowycz2022solving, azerbayev2023llemma} and code~\citep{le2022coderl, chae2024language}. Building on this, recent work proposes ``test-time scaling,'' which further lengthens the chain-of-thought \citep{snell2024scaling, 
muennighoff2025s1simpletesttimescaling}. While such methods have proven effective for puzzles like Sudoku~\citep{Sudoku-RWKV} and Hex~\citep{jones2021scaling}, where action spaces are limited, mathematical reasoning remains relatively unexplored, largely due to its exponentially larger search space. To address this challenge, researchers have investigated external verifiers—such as best-of-N selection~\citep{wang2022self}, Monte Carlo Tree Search~\citep{guan2025rstar, tian2024toward, feng2023alphazero}, and process/outcome reward modeling~\citep{zhang2025lessons, liu2025prime}. Meanwhile, state-of-the-art LLMs~\citep{openai2024o1, openai2024o3mini} are capable of self-correction—often referred to as ``system 2'' reasoning~\citep{xiang2025towards}—without explicit external verification. While longer chains of reasoning provide more room for in-depth thinking, they may also amplify the risk of error propagation~\citep{bengio2015scheduled, arora-etal-2022-exposure, holtzman2019curious}, making them more susceptible to out-of-domain disturbances such as language variation~\citep{zhao2023large, chen2024not}. From this vein, it remains unclear whether these strategies robustly generalize to new questions~\citep{matharena}, languages, or domains.

In this work, we investigate the linguistic generalizability of test-time scaling methods by proposing a fine-grained multilingual complex reasoning benchmark, showing that test-time scaling alone \emph{does not} yield robust multilingual performance. We build MCLM (\textbf{M}ultilingual \textbf{C}ompetition \textbf{L}evel \textbf{M}ath), a math reasoning dataset composed of four subsets varying source covering 55 languages.

We analyze three test-time scaling methods, outcome reward modeling \citep[ORM]{wang2022self}, process reward modeling \citep[PRM]{zhang2025lessons}, and budget forcing \citep[BF]{muennighoff2025s1simpletesttimescaling}. We examine (1) \textit{accuracy} to determine whether models retain overall performance across languages and (2) \textit{consistency} to observe whether models can solve the same questions in different languages. While \textbf{ORM} and \textbf{PRM} provide clear gains on relatively easy datasets, the improvements are marginal for challenging tasks and inconsistent across the languages. Meantime, \textbf{BF} delivers noticeable gains only in English for tougher questions, with minimal impact on other languages. These findings underscore that while test-time scaling can enhance accuracy under certain conditions, it does not guarantee robust or consistent performance across multiple languages.

Finally, we introduce \textbf{MR1-1.5B}, an open multilingual thinking LLM trained on Deepseek-R1-1.5B using 100k R1-distilled instances translated by GPT-4o. Despite having only 1.5B parameters, \textbf{MR1} achieves performance on par with GPT-4o-Mini in multilingual mathematical reasoning.

\section{Multilingual Competition Level Math}

In this section, we introduce Multilingual Competition Level Math (\textbf{MCLM}), a multilingual math reasoning benchmark with challenging competition-level questions in 55 languages.

\paragraph{Going beyond math word problems} A translated version of GSM8K~\citep{cobbe2021training}, MGSM~\citep{shi2022language}, has been widely used to assess the mathematical reasoning skills of multilingual LLMs \citep{anil2023palm2technicalreport, shao2024deepseekmathpushinglimitsmathematical, aryabumi2024aya23openweight}. However, in Table~\ref{tab_mgsm}, we observe that recent LLMs saturate MGSM. This implies the limitations of simple math word problems in accurately assessing the math reasoning capabilities of LLMs and necessitates a higher degree of complexity in reasoning benchmarks.

\paragraph{Assessing complex reasoning capabilities} In this vein, recent studies have evaluated the reasoning capabilities of LLMs using competition-level math questions \citep{AIME2024, gao2024omni}. While these benchmarks address limitations in simple math word problems, they are largely restricted to English and Chinese, limiting their ability to study multilingualism at scale.


\begin{table}[t!]
    \centering
    \fontsize{9}{11}\selectfont
    \begin{tabular}{lc}
    \toprule
    \textbf{Models} & \textbf{MGSM} \\
    \midrule
    Gemma2-9B & 78.37 \\
    Qwen2.5-14B-Instruct & 82.27 \\
    \midrule
    Qwen2.5-72B-Instruct & 88.16 \\
    Mistral-Large & 89.01 \\ 
    GPT-4o-mini & 87.36 \\
    o3-mini & \textbf{89.30} \\ 
    \bottomrule
    \end{tabular}
    \caption{\footnotesize \textbf{MGSM performance of different models.} The 2025-01-31 version is used for o3-mini, remaining scores were sourced from the \citet{yang2024qwen2}.}
    \label{tab_mgsm}
\end{table}

\subsection{Curating the MCLM benchmark}\label{mclm_benchmark}

\begin{table*}[ht]
\centering
\fontsize{9}{11}\selectfont
\begin{tabular}{@{}l l c c r@{}}
\toprule
\textbf{Subset}    & \textbf{Source Benchmark} & \textbf{Languages} & \textbf{Sample Size per Language} & \textbf{Evaluation Method} \\ 
\midrule
MT-MATH100  & Math-500   & 55  & 100 & Rule-based verifier \\
MT-AIME2024 & AIME 2024  & 55  & 30  & Rule-based verifier \\
M-IMO       & IMO (2006, 2024)      & 38  & 22--27 & LLM-as-a-Judge \\
M-MO        & Domestic/Regional Olympiads & 11 & 28--31 & LLM-as-a-Judge \\
\bottomrule
\end{tabular}
\caption{\footnotesize \textbf{Overview of benchmark subsets}: source benchmarks, language coverage (full lists in the appendix), sample sizes, and evaluation methods.  Please see Appendix~\ref{app:languages} for the full list of languages.}
\label{tab_euler_bench}
\end{table*}


\paragraph{Machine-translated reasoning} We select AIME and MATH-500 \citep{lightman2023let}, two widely used mathematical benchmarks, as the main source of complex math questions. For 100 questions randomly sampled from MATH-500 and full AIME datasets, we translate both benchmarks with GPT-4o \citep{openai2024gpt4ocard}, as shown to be proficient in translating mathematical contexts~\citep{chen-etal-2024-breaking, lai2023okapi}. We then verified that the answers and equations remained unchanged after translation, removing one sample from MATH500 due to translation inconsistencies. Both subsets consist of questions with numerical answers only. For further details on the machine translation and sampling process, see Appendix~\ref{app:sampling_math100}.


\paragraph{Human-annotated reasoning} To mitigate the potential biases in machine-translated data from translation artifacts~\citep{plaza2024spanish, son2024kmmlu}, we also include human-translated or originally written questions. 

First, we manually review 114 International Mathematical Olympiad \citep[IMO]{IMO} questions from 2006 to 2024 in English, excluding proof-based and image-heavy problems, resulting in a final set of 27. We then collect their official translations in 38 languages. Where official translations are unavailable, we do not substitute machine-generated versions, leaving those entries missing.

Second, we gather problems from various domestic and regional mathematical Olympiads worldwide. These contests originate in multiple languages, providing valuable data for multilingual mathematical reasoning. For English, Chinese, and Korean—where competition-level benchmarks already exist \citep{he2024olympiadbench, ko2025understand}—we incorporate existing datasets rather than recollecting data. While we exclude proof-based questions for simplicity, the final dataset still features a diverse range of answer formats (e.g., numerical, Boolean, descriptive) and spans 11 languages. We use GPT-4o-mini\footnote{2024-07-18 version \citep{openai2024gpt4ocard}} for evaluation. An overview is provided in Table~\ref{tab_euler_bench}, additional details are in Appendix~\ref{app:mimo_mmo}.

\section{Experimental Settings}

 In this section, we provide an overview of the test-time scaling methods evaluated (Section~\ref{sub_sec_tts}), compare the inference budgets in terms of FLOPs across different scaling techniques (Section~\ref{sub_sec_budget}), and describe the evaluation metrics used to assess their performance (Section~\ref{sub_sec_metrics}). 

\begin{figure}[t]
\centering
\includegraphics[width=\columnwidth]{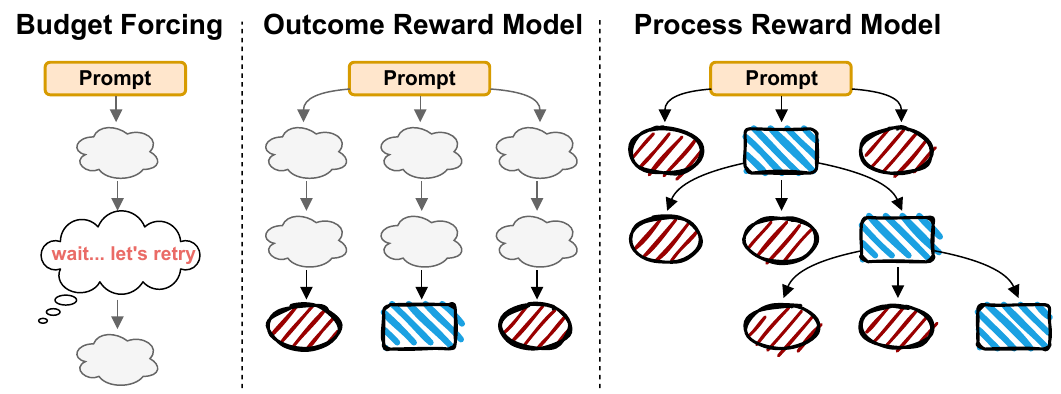}
\caption{\footnotesize \textbf{Comparison of different inference-time scaling strategies.} Blue boxes represent selected outputs, while red boxes indicate rejected ones.}
\label{fig_inference}
\end{figure}

\subsection{Baselines: Test-Time Scaling Strategies}\label{sub_sec_tts}
In this work, we evaluate three test-time scaling strategies using Qwen2.5-1.5B and 7B instruct models~\citep{yang2024qwen2O} as baselines (Figure~\ref{fig_inference}). We selected these model sizes because they offer a balanced trade-off between reasoning capacity and computational efficiency. Models smaller than 1.5B lack the capacity to solve complex problems, while larger models can be prohibitively expensive to scale~\citep{biderman2023pythia}.

\paragraph{Outcome Reward Modeling}~We generate \(N\) responses per instance and use Qwen2.5-Math-72B-RM~\citep{yang2024qwen2} to evaluate them, selecting the highest-scoring answer as the final output.

\paragraph{Process Reward Modeling}~In contrast to outcome reward modeling, this strategy integrates the reward model during inference to guide the generation process. We employ Qwen2.5-Math-72B-PRM~\citep{zhang2025lessons}; the model generates \(c\) candidate continuations at each step and selects the best one. For both ORM and PRM, the generator and reward model are served on separate servers, thereby avoiding the overhead of repeatedly on- and off-loading model weights.

\paragraph{Budget Forcing}~Recent LLMs, such as R1~\citep{guo2025deepseek} and O1~\citep{openai2024o1}, are designed to generate longer chain-of-thoughts with in-context exploration and correction, allowing them to naturally scale during inference. However, this approach lacks controllability. To mitigate this, we adopt the budget-forcing method proposed by \citet{muennighoff2025s1simpletesttimescaling}. In budget forcing, these thinking models are truncated and required to output an answer if they exceed a predefined budget. Conversely, if they fall short of the budget, they are prompted to generate additional reasoning steps, encouraging further exploration and correction.

\subsection{Calculating Inference FLOPs}\label{sub_sec_budget}

\begin{figure}[t!]
\centering
\includegraphics[width=\columnwidth]{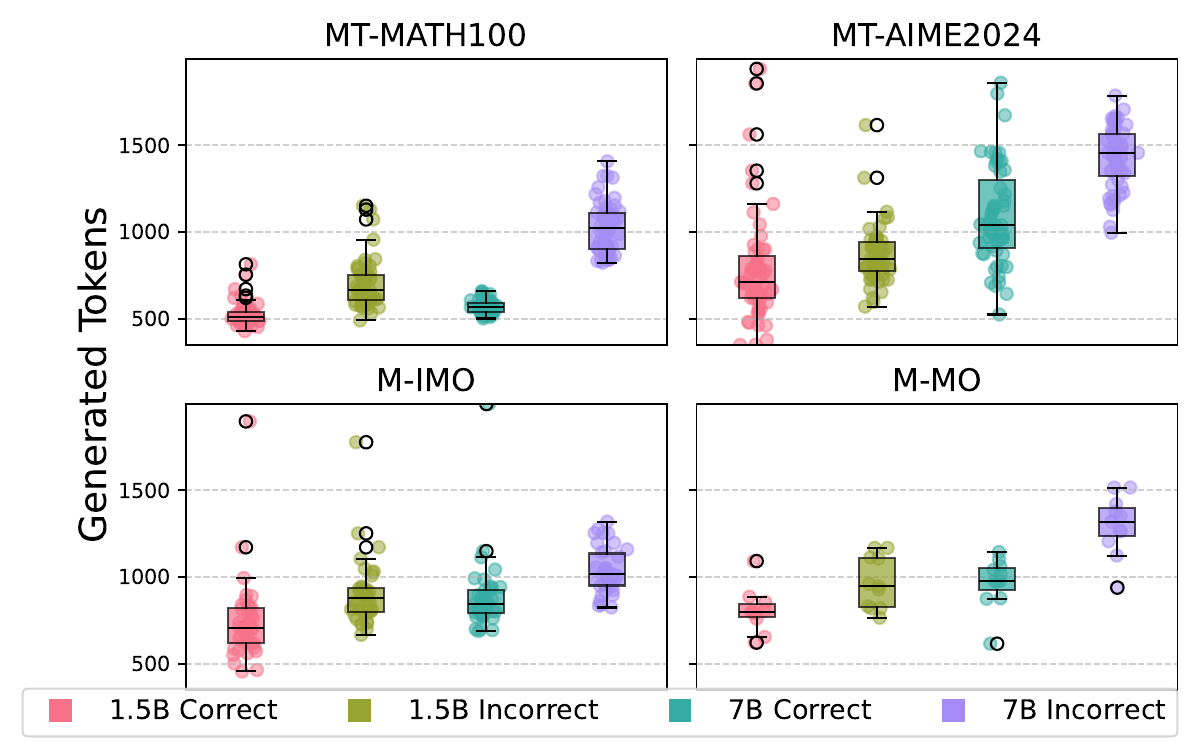}
\caption{\footnotesize \textbf{\# of generated tokens for 1.5B and 7B models in a greedy setting, divided by correctness.} Languages are represented as scatter plots, overlaid on box plots.}
\label{fig_tokens}
\end{figure}

For our experiments, we first establish a unified inference budget based on two key estimates: the generator’s cost is approximated as \(2N_GD\)~\citep{kaplan2020scaling}, where \(N_G\) is the number of parameters in the generation model and \(D\) is the total number of tokens generated per instance. The verifier’s (or reward model’s) cost is estimated as \(4N_V\), where \(N_V\) is the number of parameters in the verifier. Here, the multiplier of 4 for the reward model’s cost reflects a base cost of \(2N_V\) that is doubled to account for the additional overhead incurred when invoking the reward model during inference~\citep{snell2024scaling}. In the ORM, we generate \(k\) responses per instance, leading to a total inference cost of 
\begin{equation}
    k \times \Bigl(2N_GD + 4N_V\Bigr).
\end{equation}


In Figure~\ref{fig_tokens}, under greedy generation, we observe that models tend to generate longer responses once they produce an error, particularly on harder benchmarks. Additionally, the 7B model generally produces longer outputs than the 1.5B model. However, no systematic trends emerge across the 55 languages—there is no clear pattern, such as longer outputs for low-resource languages. Although token counts vary with configuration, these differences are negligible compared to the dominant effect of model size on inference FLOPs. Consequently, we use an average of 921 tokens per question to estimate cost.
\paragraph{Outcome Reward Modeling} In ORM with \(k=2\) responses, the inference cost is approximated as
\begin{equation}
    \text{ORM FLOPs} \approx 2\Bigl(2N_G \times 921 + 4N_V\Bigr).
\end{equation}
Assuming \(N_G = 1.5\times10^9\) and \(N_V = 72\times10^9\), this configuration results in an estimated cost of approximately \(6.10\times10^{12}\) FLOPs per instance.

\paragraph{Process Reward Modeling} In PRM, at each generation step, the model produces \( c \) candidates, with each candidate generating \( x \) tokens (we fix \( x = 128 \) in our experiments). The total inference cost over \( S \) steps is given by
\begin{equation}
    \text{PRM FLOPs} = S\,c\Bigl( x \cdot 2N_G + 4N_V \Bigr).
\end{equation}

For PRM configurations, \(S\) and \(c\), our preliminary experiments indicated that scaling one parameter in isolation produced suboptimal performance: a high \(S\) with a low \(c\) failed to explore sufficient alternatives, while an excessively low \(S\) prevented the generation process from completing. Therefore, we opted to proportionally scale both \(S\) and \(c\) to achieve a balanced search during generation.

\paragraph{Budget Forcing} In contrast, BF relies solely on the generator, so its inference cost is given by
\begin{equation}
    \text{SC FLOPs} = 2N_G \cdot BF,
\end{equation}
where \( BF \) denotes the effective number of tokens that may be generated during the inference with budget forcing. In Table~\ref{tab:config_sc}, we select the PRM parameters \( S \) and \( c \) and adjust the BF token budget \( BF \) so that the overall inference cost of each method matches that of ORM for \( k = 2, 4, \) and \( 8 \).

\begin{table}[t!]
\centering
\fontsize{9}{11}\selectfont
\begin{tabular}{ccc}
\toprule
\( k \) &  \((S, c)\) & \( BF \) \\
\midrule
2 & (3, 3) & \(\approx 2048\) tokens \\
4 & (4, 5) & \(\approx 4096\) tokens \\
8 & (5, 8) & \(\approx 8192\) tokens \\
\bottomrule
\end{tabular}
\caption{\footnotesize \textbf{Selected configurations for PRM and BF}. Each \(S\), \(c\), and \( BF\) is set so that the inference FLOPs match ORM.}
\label{tab:config_sc}
\end{table}

\subsection{Evaluation Metrics}\label{sub_sec_metrics}

We evaluate our models using two primary metrics that capture performance at multiple levels: (1) \textit{accuracy} and (2) \textit{cross-lingual consistency}.

\paragraph{Accuracy}
We measure accuracy at a surface level to determine whether a single model achieves comparable performance across different languages.

\paragraph{Cross-Lingual Consistency}
To examine whether the model tends to solve (or fail) the \emph{same} questions across languages, we compute \textbf{Fleiss’ kappa} \citep{fleiss1971measuring}, which is originally designed to measure agreement among multiple annotators. In our setup, however, we treat each language as an ``annotator’’: for each problem, each ``annotator’’ (i.e., each language version of the model) provides either a correct or incorrect label. We then define consistency through Fleiss' kappa as:
\begin{gather}
\kappa = \frac{\bar{P} - \bar{P}_e}{1 - \bar{P}_e}\\
\bar{P} = \frac{1}{N}\sum_{i=1}^N \frac{1}{n(n-1)}\sum_{j=1}^k n_{ij}(n_{ij}-1)\\
\bar{P}_e = \sum_{j=1}^k p_j^2, \quad p_j = \frac{1}{Nn}\sum_{i=1}^N n_{ij},
\end{gather}
where \(N\) is the number of problems, \(n\) is the number of languages, and \(n_{ij}\) is the count of how many times language \(j\) gives a particular label (correct or incorrect) for problem \(i\). In this formulation, \(\bar{P}\) is the observed agreement (i.e., the proportion of problems for which all languages concur on correctness or incorrectness), and \(\bar{P}_e\) is the expected agreement by chance. A high Fleiss’ kappa indicates that the model responds consistently across languages (solving the same problems), not merely achieving similar overall accuracy by chance.


\section{Result 1: ORM and PRM}

In this section, we assess the multilingual robustness of both ORM and PRM. We find that, while each approach can boost performance, these gains do not consistently generalize across different languages and levels of difficulty.

\subsection{Outcome Reward Modeling}

\begin{figure}[t!]
\centering
\includegraphics[width=\columnwidth]{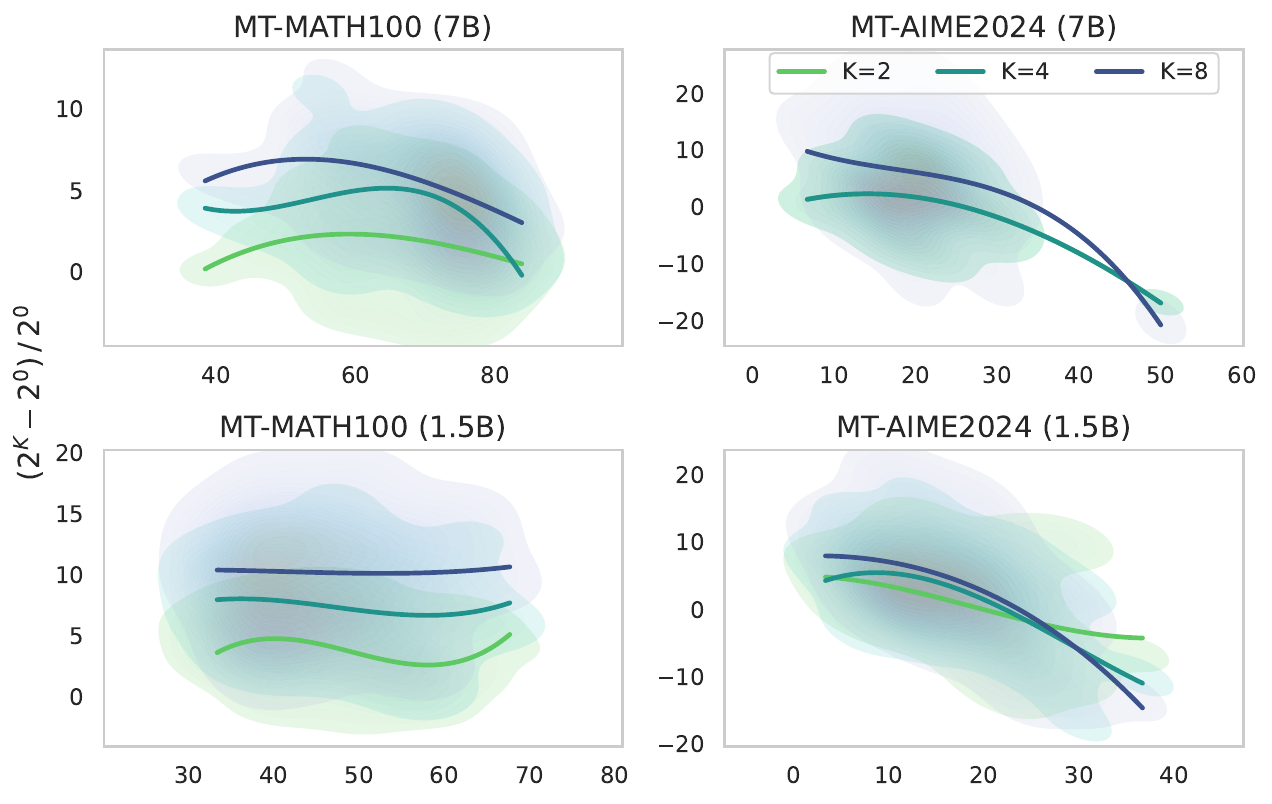}
\caption{\footnotesize \textbf{Gains of ORM compared to a greedy-decoding baseline.} The semi-transparent “cloud” indicates the 2D data distribution via a KDE density plot, and the overlaid lines are third-order polynomial regressions modeling how each ORM setting scales with the baseline score.}
\label{fig_orm_results}
\end{figure}

For ORM, we use Qwen2.5-Math-1.5B and 7B-Instruct models to generate \(K\) samples per query and then apply Qwen2.5-Math-72B to score each sample, selecting the one with the highest score. 

\paragraph{Limited gains at scale in non-English settings}~In Figure~\ref{fig_orm_results}, we plot each model’s baseline performance (averaged across 55 languages) on the \(x\)-axis versus the relative gain of each ORM setting (with \(K \in \{2,4,8\}\)) on the \(y\)-axis. On the MT-MATH100 dataset, both the 1.5B and 7B models show consistent improvement as \(K\) increases. However, on the more challenging MT-AIME2024 dataset, the gains for different \(K\) values are largely indistinguishable and, in some cases, even negative. This trend is comparable to English, which shows steady improvements also on MT-AIME2024—for instance, the 1.5B model rises from 16.67 to 26.67 to 36.67 as \(K\) increases, while the 7B model goes from 20.00 to 26.67 to 36.67.

Overall, while ORM is a viable scaling strategy in English, it yields limited returns in many other languages—possibly due to the models' difficulty in generating high-quality candidates. With few plausible options available, the reward model cannot effectively identify an improved solution.

\subsection{Process Reward Modeling}

\begin{figure}[t!]
\centering
\includegraphics[width=\columnwidth]{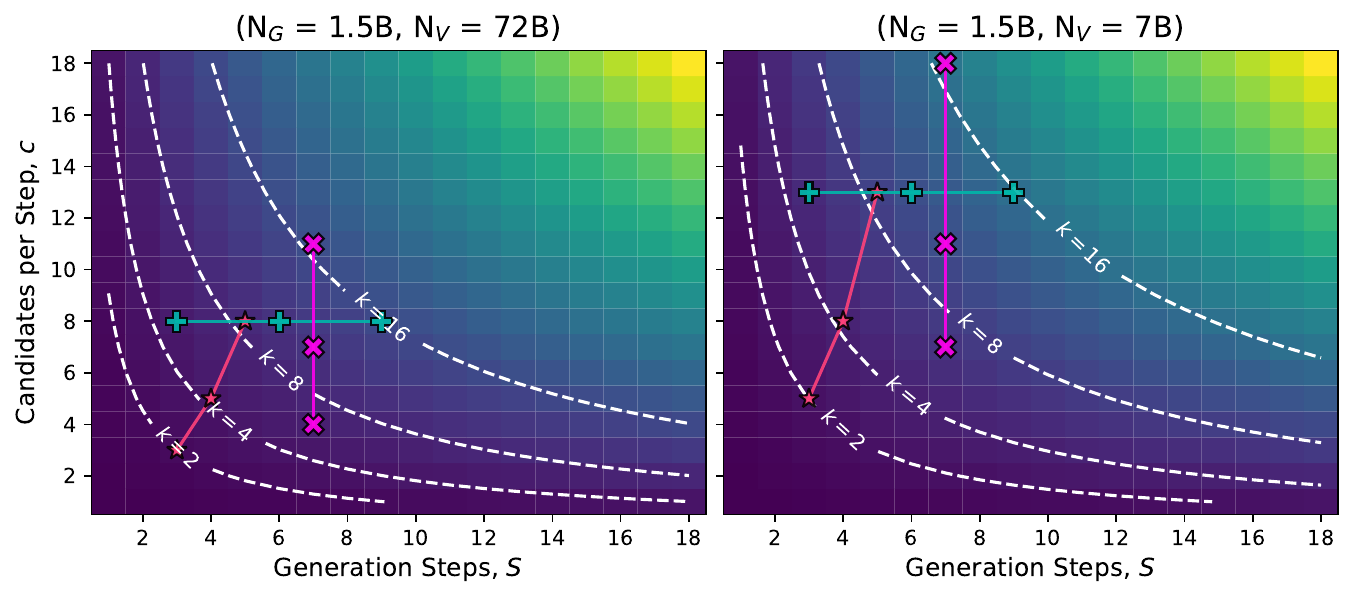}
\caption{\footnotesize \textbf{PRM inference FLOPs as a function of generation steps $S$ and candidates per step $c$.} The left panel uses a verifier size of 72B, while the right panel uses a 7B RM, displaying adjusted configurations to yield similar costs.}
\vspace{-3pt}
\label{fig_prm_configs}
\end{figure}

Along with the configurations mentioned in Table~\ref{tab:config_sc}, we experiment with additional setups to study how PRMs scale. Details are provided in Figure~\ref{fig_prm_configs}. In general, we evaluate two approaches: fixing \(S\) while increasing \(C\) (pink) and fixing \(C\) while scaling \(S\) (green). Additionally, we compare the efficacy of a 7B verifier against the original 72B. Due to cost constraints, these configurations are tested only on 14 languages of MT-MATH100 using Qwen2.5-Math-1.5B-Instruct.

\begin{figure*}[t!]
\centering
\includegraphics[width=\textwidth]{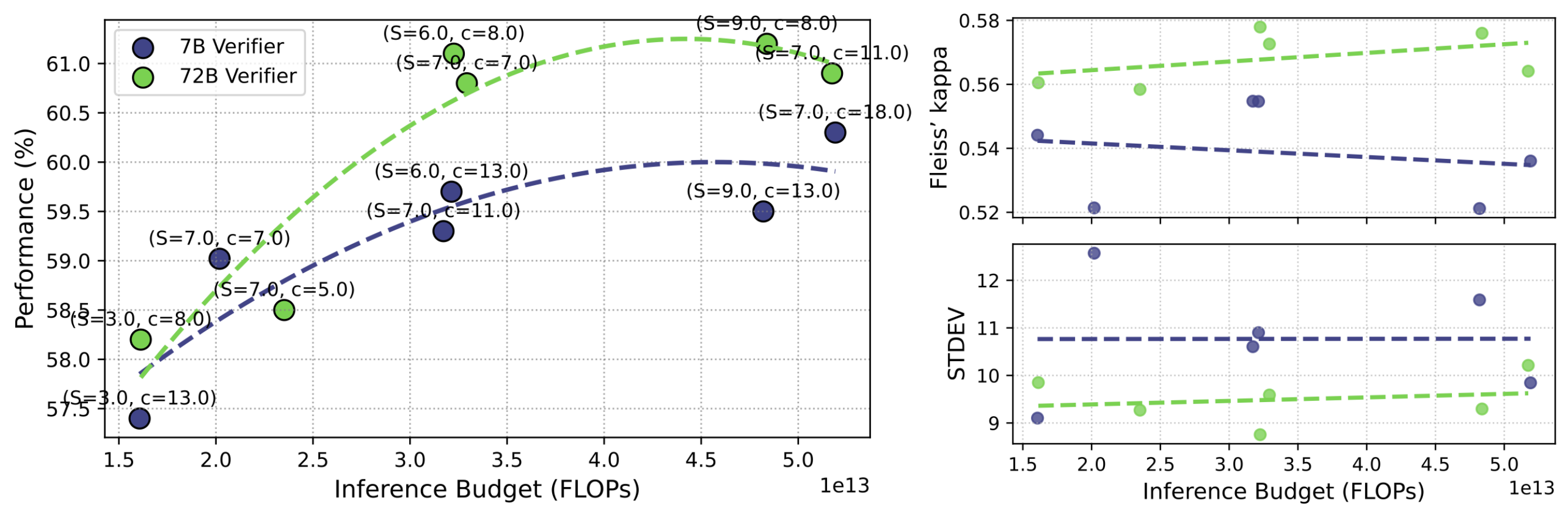}
\caption{\footnotesize \textbf{Inference FLOPs versus PRM performance and consistency.} (Left) Second‐degree polynomial regressions for average performance on 14 languages, comparing the 7B (blue) and 72B (green) reward models. (Right) Fleiss’ kappa (top) and standard deviation (bottom) plotted against the same FLOPs budget; the fitted curves reveal no clear monotonic trend.}
\label{fig_prm_results}
\end{figure*}

\paragraph{No scalable gains for variance or consistency} Figure~\ref{fig_prm_results} shows that, in PRM, the average performance of Qwen2.5-Math-1.5B-Instruct increases steadily with the inference budget. Even though the 7B reward model provides a larger search space, the 72B reward model achieves better outcomes under comparable compute. From a hardware standpoint, it can be more effective to run fewer steps and rely on a larger verifier. However, no clear pattern emerges for Fleiss’ kappa or the standard deviation of individual scores, suggesting that adding more budget does not necessarily improve model consistency or reduce variance. In practical terms, while accuracy may scale with compute for PRM, cross-lingual consistency does not appear to follow, even for a relatively easier dataset like MT-MATH100.



\subsection{ORM over PRM}

\begin{figure}[t!]
\centering
\includegraphics[width=\columnwidth]{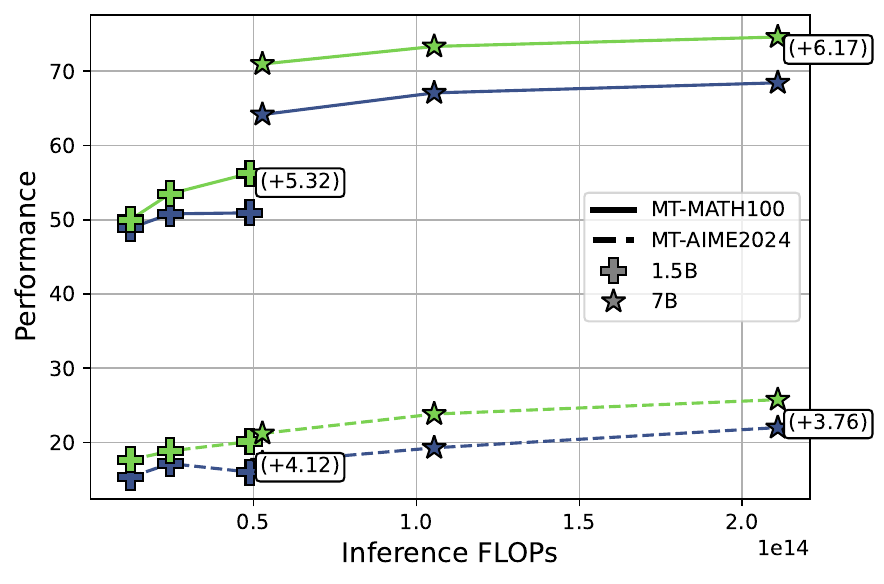}
\caption{\footnotesize \textbf{Comparison of PRM vs. ORM performance on MATH (solid lines) and AIME (dashed lines).} 1.5B models are shown with plus markers, 7B models with stars. Blue lines represent PRM, green lines represent ORM. White box annotations indicate the performance difference (ORM − PRM) at the highest compute setting for each line.}
\label{fig_orm_prm}
\end{figure}

As discussed earlier, both ORM and PRM exhibit unstable multilingual performance growth, with greater variance and lower Fleiss' Kappa scores at higher inference FLOPs. However, despite this instability, as shown in Figure~\ref{fig_orm_prm},  ORM consistently outperforms PRM in average accuracy, suggesting that, in general, it may be the more reliable choice. This is especially true given that, despite being assigned the same inference FLOPs, PRM invokes the reward model more frequently, requiring iterative back-and-forth interactions between the generator and evaluator, leading to higher latency.

\section{Result 2: Budget Forcing}\label{budget_forcing}

LLMs with “system 2” reasoning~\citep{xiang2025towards}—such as OpenAI’s newest O-series models~\citep{openai2024o1}, Google’s Gemini Thinking\footnote{\url{https://deepmind.google/technologies/gemini/flash-thinking/}}, and Deepseek R1~\citep{guo2025deepseek}—are emerging as a test-time scaling approach. By generating and refining responses within a single inference (without external verifiers), these models seek to dramatically expand the inference budget for improved performance. In this section, we examine their effectiveness as a test-time scaling strategy. Because proprietary solutions remain largely opaque, we train our own LLMs with system 2 thinking. Here, we describe our training methods (Section~\ref{sub_sec_train}), report performance on MCLM (Section~\ref{sub_sec_perf}) and scaling affects from budget forcing(Section~\ref{sub_sec_scale}).

\begin{table*}[t!]
\centering
\fontsize{9}{11}\selectfont
\begin{tabular}{l|cccccc}
\toprule
Models & MT-MATH100 & MT-AIME2024 & M-IMO & M-MO & Average \\
\midrule
Qwen2.5-Math-1.5B-Instruct & 42.32 $\pm$ 8.61  & 16.36 $\pm$ 6.89  & 12.23 $\pm$ 6.02  & 25.00 $\pm$ 19.10  & 23.98 \\
Deepseek-R1-1.5B          & 49.40 $\pm$ 8.84  & 17.21 $\pm$ 6.69  & 21.94 $\pm$ 6.75  & 26.77 $\pm$ 19.83  & 28.83 \\
GPT-4o-Mini               & 70.30 $\pm$ 3.68  & 20.18 $\pm$ 6.83  & 13.33 $\pm$ 5.36  & 30.81 $\pm$ 15.80  & 33.66 \\
o3-Mini                  & \textbf{84.89} $\pm$ 2.80  & \textbf{45.33} $\pm$ 5.35  & \textbf{29.75} $\pm$ 6.86  & \textbf{51.42} $\pm$ 16.94  & \textbf{52.85} \\
\midrule
Qwen2.5-Math-1.5B + SFT    & 37.47 $\pm$ 7.56  & 14.85 $\pm$ 6.69  & 10.50 $\pm$ 5.16  & 18.40 $\pm$ 14.92  & 20.30 \\
Qwen2.5-Math-1.5B + MT-SFT & 42.02 $\pm$ 7.46  & 16.67 $\pm$ 7.31 &  10.52 $\pm$ 4.63  & 19.92 $\pm$ 12.68 & 22.28   \\
Deepseek-R1-1.5B + MT-SFT  & \textbf{55.61} $\pm$ 10.93 & \textbf{19.94} $\pm$ 8.10  & \textbf{19.20} $\pm$ 6.24  & \textbf{28.97} $\pm$ 16.64  & \textbf{30.93} \\
\bottomrule
\end{tabular}
\caption{\footnotesize \textbf{Model performance across MCLM.} Best model highlighted in \textbf{bold} for each panel. For results per language see Appendix~\ref{app_add_results}.}
\label{tab:performance}
\end{table*}

\subsection{Inducing System 2 Thinking}\label{sub_sec_train}
   
A number of concurrent works propose diverse strategies for developing LLMs with long thinking. Broadly, these approaches fall into two main categories: (1) online reinforcement learning with verifiable outputs \citep{deepscaler2025, ye2025emergence}, and (2) supervised fine-tuning on the "thinking trajectories" of proprietary LLMs \citep{muennighoff2025s1simpletesttimescaling, ye2025limo}. However, \citet{deepscaler2025} reports requiring 3,800 A100 GPU hours to induce such behavior, making it prohibitively expensive for our setting. Instead, we opt for supervised fine-tuning using thinking trajectories distilled from R1. Below are three training configurations.

    
\paragraph{Qwen2.5-Math-1.5B + SFT}~Following \citet{muennighoff2025s1simpletesttimescaling} and  \citet{huang2024o1}, we randomly sample 50K thinking trajectories generated by R1 from the OpenR1-220K~\footnote{\url{https://huggingface.co/datasets/open-r1}} dataset and fine-tune our model for three epochs.

\paragraph{Qwen2.5-Math-1.5B + SFT with Translated Data}~While training exclusively on English math problems has been shown to be effective and can generalize to new languages to some extent~\citep{liu2024acemath}—likely due to the universal nature of mathematical logic—we explore whether translating the data helps training. Following \citet{ko2025understand, zhang2023plug}, we translate the problem and solution components into 14 languages\footnote{See Appendix~\ref{app:languages} for the complete list of languages.} using GPT-4o, while keeping the reasoning process in English to leverage the model’s strong proficiency in English-based logical reasoning. For further details on the training dataset, see Appendix~\ref{app_training}.

\paragraph{Deepseek-R1-1.5B + SFT with Translated Data}~Finally, we try initializing the translated SFT from Deepseek-R1-1.5B (hereafter MR1-1.5B). Since the model is already proficient in generating extended reasoning, we observe that longer training leads to performance degradation. To mitigate this, we terminate training at 0.5 epochs.

\begin{figure}[t!]
\centering
\includegraphics[width=\columnwidth]{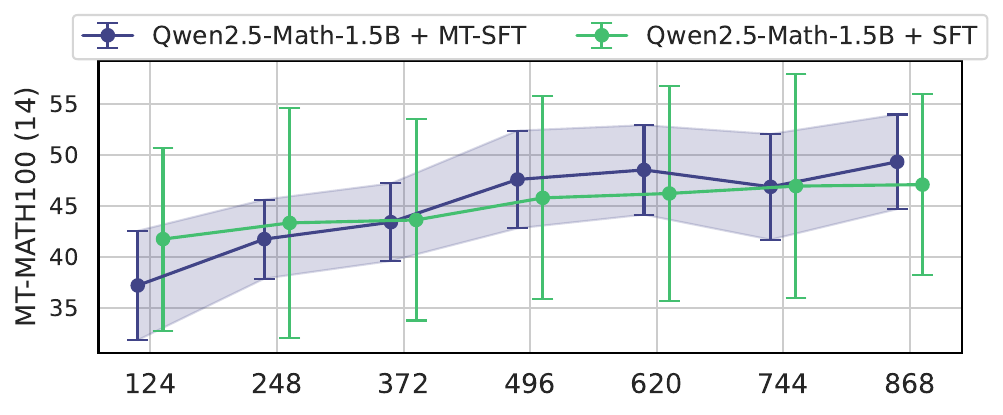}
\caption{\footnotesize \textbf{Performance of Qwen2.5-Math-1.5B +SFT and + MT-SFT at each training checkpoint.} Average score and error bars for each checkpoint are displayed. The shaded region is the mean $\pm$ standard deviation for MT-SFT.}
\label{fig_training}
\end{figure}

\subsection{Performance of trained models}\label{sub_sec_perf}

\paragraph{Translated data improves cross-lingual performance.}Table~\ref{tab:performance} compares Qwen2.5-Math-1.5B and Deepseek-R1-1.5B under various fine-tuning regimes, revealing two key trends. First, incorporating translated data into Qwen2.5-Math-1.5B delivers a modest +1.98\% improvement over an English-only setup, indicating that relying exclusively on English data is insufficient for robust cross-lingual performance \citep{liu2024acemath}. As shown in Figure~\ref{fig_training}, models trained on multilingual inputs begin with lower accuracy—likely due to increased entropy—but soon surpass their English-only counterparts and maintain lower variance across languages. Second, initiating fine-tuning from Deepseek-R1-1.5B, already adept at extended chain-of-thought reasoning, yields even greater gains on MT-AIME2024, M-IMO, and M-MO, performing on par with GPT-4o-Mini. Notably, MR1-1.5B reaches an average score of 30.93 (+2.1\% over its original baseline) with just 0.5 epochs of training, underscoring how a self-correcting model more readily benefits from multilingual data. Collectively, these results suggest that while incorporating translated data benefits a monolingual base, leveraging a model with established self-correction capabilities can amplify these gains in multilingual math reasoning.

\subsection{Budget-Constrained Scaling}\label{sub_sec_scale}

\begin{figure}[t!]
\centering
\includegraphics[width=\columnwidth]{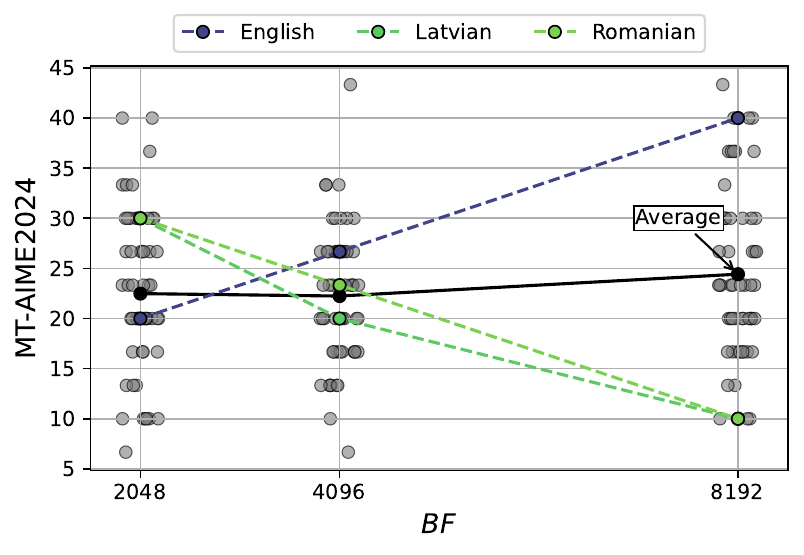}
\caption{\footnotesize \textbf{Performance of MR1 on MT-AIME2024 at \(BF = \{2048, 4096, 8192\}\).} Grey dots represent individual languages. Solid lines indicate average performance, while dashed lines highlight reference performances for selected languages.}
\label{fig_budget_forcing}
\end{figure}

To better compare self-correction with other scaling methods (e.g., ORM and PRM), we examine its performance under fixed inference budgets. We apply the budget forcing approach introduced by \citet{muennighoff2025s1simpletesttimescaling} to constrain the generation budget of MR1-1.5B. Following the budget settings in Table~\ref{tab:config_sc}, during inference, if the model reaches 90\% of its allocated budget, we truncate the output and append "The final answer is" to prompt a concise answer. Conversely, if the model completes generation before reaching the limit, we truncate at the last line break, append "Wait...", and prompt the model to continue generating.

\paragraph{R1-like LLMs offer no clear edge over ORM}~Contrary to the recent surge of interest in ``system 2'' LLMs, that scale test-time compute by generating long reasoning traces, constraining these models to the same inference budgets reveals no clear advantage over test-time scaling methods such as ORM or PRM (see Figure~\ref{fig_title}). As shown in Figure~\ref{fig_budget_forcing}, budget forcing yields nearly linear performance gains for English but provides limited benefits for most other languages. The distribution remains largely unchanged, achieving an overall average increase of only 1.9\% as \(BF\) scales from 2048 to 8096. In some cases, such as Latvian and Romanian, performance even declines. Implying that there is scant evidence that the variance in performance diminishes or that coverage expands uniformly.

\section{Related Works}

\paragraph{Test-Time Scaling}~As concerns grow that the benefits of scaling pre-training compute may be saturating~\citep{longpre2024consent}, research has shifted toward \emph{test-time scaling}, which expands the notion of chain-of-thought reasoning~\citep{wei2022chain}. Intuitively, the reasoning capacity of an LLM is limited by the number of tokens it can generate; hence, more challenging questions may require a longer chain of thought~\citep{wu2025more}. An early example is self-consistency CoT~\citep{wang2022self}, which generates multiple responses and selects the best via voting. This idea has since been developed into more cost-effective strategies for searching broader solution spaces (e.g., tree-of-thought methods~\citep{yao2024tree}, Monte Carlo Tree Search~\citep{guan2025rstar}, and process supervision~\citep{zhang2025lessons, luo2024improve}). Recently, models trained with online reinforcement learning~\citep{shao2024deepseekmathpushinglimitsmathematical} appear to exhibit an ``aha moment,''~\citep{guo2025deepseek} wherein they dynamically decide to generate longer sequences to iteratively explore, solve, and self-correct.

\paragraph{Mathematical Reasoning in Non-English}~Early attempts at multilingual math reasoning involved supervised fine-tuning on translated datasets~\citep{chen2023breaking, lai2024mcot}, but performance often deteriorated when models shifted away from their original language embeddings~\citep{hong2024cross}. To minimize such degradation, more recent work has increasingly relied on English as a pivot language. This approach can be implemented in various ways: either internally, by mapping multilingual inputs into an English-centric latent space \citep{yoon2024langbridge, fan2025slam, zhu2024question, she2024mapo}, or externally, by translating non-English tasks into English and then back to the target language \citep{zhang2023plug, ko2025understand}. Although this strategy has reduced the performance gap between English and other languages, the stability of transfer under different training conditions remains underexplored. Moreover, many studies rely on the MGSM benchmark~\citep{shi2022language}, which appears too easy for large-scale models or those enhanced by advanced reasoning techniques such as test-time scaling.

\section{Conclusion}

In this work, we examine the linguistic generalizability of three test-time scaling methodologies—Outcome Reward Modeling (ORM), Process Reward Modeling (PRM), and Budget Forcing (BF)—under budget-constrained settings. Using Qwen2.5-1.5B Math as a generator, ORM achieves a 35.84 score on our newly introduced multilingual math benchmark, \textbf{MCLM}, which spans 55 languages. With BF, \textbf{MR1-1.5B}—our multilingual LLM demonstrating extended reasoning—attains 35.23. Notably, once constrained to similar inference budgets, all three scaling methods exhibit comparable levels of improvement. Additionally, although these approaches appear promising in English (e.g., a 20-point improvement on AIME for the 1.5B model), we find that such gains do not consistently extend to other languages, where improvements average only 1.94 points—a pattern observed across all three methods. Moreover, increasing test-time compute often amplifies performance variance and reduces cross-linguistic consistency. To enable further study of these issues, we release both MCLM and MR1-1.5B.


\section*{Limitations}

Although this work focuses solely on mathematical tasks, the lack of multilingual generalization we observe could be even more pronounced in areas requiring extensive cultural or domain-specific understanding; we leave this for future works. Additionally, due to budget constraints, this work primarily focuses on smaller-scale experiments (e.g., Qwen2.5-Math-1.5B-Instruct and occasionally Qwen2.5-Math-7B-Instruct). Although these parameter ranges are commonly used in both industry and academia, the observed lack of multilingual generalization for test-time scaling may not necessarily extend to significantly larger models (\(70\text{B}\) or more). Moreover, our test-time scaling experiments also remain on the smaller side; for instance, \citet{el2025competitive} scale to as many as 1162 candidates in their best-of-\(n\) setting. (It should still be noted that even experiments at this scale required over \textbf{2500 A100 GPU hours}) Given that the so-called ``curse of multilinguality''~\citep{conneau-etal-2020-unsupervised} naturally disappears as pre-training compute grows by several orders of magnitude~\citep{aryabumi2024aya23}, it is plausible that larger models may behave differently. Nevertheless, our findings at smaller scales—where test-time scaling shows no indication of fostering robust multilingualism—remain valuable, as they reveal potential boundaries for less resource-rich setups and highlight directions for future research.


\bibliography{custom}

\appendix

\newpage

\onecolumn

\section{Additional details on MCLM}
In this section, we provide additional details on the MCLM benchmark, including the languages covered by each subset (Section~\ref{app:languages}), the sampling process for MT-MATH100 (Section~\ref{app:sampling_math100}), the sources for the M-IMO, and M-MO subset (Section~\ref{app:mimo_mmo}), the prompts used for GPT-4o and -mini (Section~\ref{app:prompts}), and contamination considerations (Section~\ref{app:contamination}).

\subsection{Details in Language Coverage}\label{app:languages}

We examine four groups of languages in this paper: (A) the 55 languages into which MATH500 and AIME2024 have been translated, (B) the 14 languages frequently sampled to reduce evaluation costs, (C) the languages covered in M-IMO, and (D) those in M-MO. The complete list for each group is provided in Table~\ref{tab:lang_appendix}.

\begin{table*}[ht]
\centering
\fontsize{9}{11}\selectfont
\begin{tabular}{@{}l|p{10cm}|c@{}}
\toprule
\textbf{Lang. Group} & \textbf{Languages (ISO Codes, Sorted Alphabetically)} & \textbf{\# Lang.} \\
\midrule

(A) & \texttt{af, ar, bg, bn, ca, cs, cy, da, de, el, en, es, et, fa, fi, fr, gu, he, hi, hr, hu, id, it, ja, kn, ko, lt, lv, mk, ml, mr, ne, nl, no, pa, pl, pt, ro, ru, sk, sl, so, sq, sv, sw, ta, te, th, tl, tr, uk, ur, vi, zh-cn, zh-tw} & 55 \\[1ex]
\midrule
(B) & \texttt{af, ar, de, en, es, fr, he, id, it, ja, ko, tr, vi, zh-cn} & 14 \\[1ex]
\midrule
(C) & \texttt{af, ar, bg, cs, da, de, el, en, et, es, fi, fr, he, hr, hu, id, it, ja, ko, lt, lv, mk, nl, no, pl, pt, ro, ru, sk, sl, sq, sv, th, tr, uk, vi, zh-cn, zh-tw} & 38 \\[1ex]
\midrule
(D) & \texttt{cs, de, en, fr, ja, ko, nl, pl, ru, sk, zh-cn} & 11 \\
\bottomrule
\end{tabular}
\caption{\footnotesize Full language lists for each dataset subset. MT-MATH100, MT-AIME2024, M-IMO, and M-MO cover 55, 38, and 11 ISO codes respectively.}
\label{tab:lang_appendix}
\end{table*}

Running evaluations on all 55 languages can be computationally intensive, particularly when testing for test-time scaling. To address this, we have created a downsampled group (B) consisting of 14 languages for our experiments. These languages were chosen to represent a broad spectrum of linguistic families and writing systems. In terms of language families, the selection includes representatives from Afro-Asiatic (Arabic, Hebrew), Austronesian (Indonesian), Japonic (Japanese), Koreanic (Korean), Turkic (Turkish), Austroasiatic (Vietnamese), and Sino-Tibetan (Chinese). Additionally, the chosen languages encompass diverse scripts: several, including Afrikaans, German, English, Spanish, French, and Italian, use the Latin alphabet; others use distinct writing systems—Arabic and Hebrew employ abjads (consonant-based scripts); Japanese combines logographic characters (Kanji) with syllabic scripts (Hiragana and Katakana); Korean is written in Hangul; Turkish uses a modified Latin alphabet; Vietnamese utilizes a Latin-based alphabet with diacritics; and Chinese is written using Chinese characters.

\subsection{Sampling MATH100}\label{app:sampling_math100}

Once creating MGSM, \citet{shi2022language} opted to randomly sample 250 questions from the GSM8K dataset for computational efficiency. Similarly, we sample 100 questions from MATH500 to keep evaluation costs manageable across multiple languages. Initially, we conduct random sampling. Before extending this approach to all 55 languages, we first apply it to language group (B). For language group (B), we create both the MT-MATH100 and MT-MATH500 versions, where entire subsets are translated for the later. We then evaluate 10 models—each trained using different methods to enhance mathematical reasoning—to determine whether the sampled MATH100 subset reliably represents the full dataset.

\begin{table*}[h]
    \centering
    \fontsize{10}{13}\selectfont
    \begin{tabular}{c|lcc|cc}
        \toprule
        \textbf{Rank} & \multicolumn{1}{c}{\textbf{Model}} & \textbf{MATH-500} & \textbf{MATH-100} & \textbf{Score Diff.} & \textbf{Rank Diff.}\\
        \midrule
        1  & o3-mini & 85.00 & 85.93 & 0.93 & - \\
        2  & Eurus-2-7B-PRIME & 73.76 & 76.63& 2.86 & - \\
        3  & Qwen2.5-Math-7B-Instruct & 73.70 & 75.98 & 2.27 & - \\
        4  & DeepSeek-R1-Distill-Qwen-32B & 72.73 & 75.98 & 3.24 & - \\
        5  & DeepSeek-R1-Distill-Qwen-7B & 67.25 & 68.69 & 1.44 & \up{1} \\
        6  & AceMath-7B-Instruct & 65.90 & 70.06 & 4.16 & \down{1} \\
        7  & AceMath-1.5B-Instruct & 65.60 & 68.19 & 2.58 & - \\
        8  & DeepSeek-R1-Distill-Qwen-1.5B & 53.74 & 56.78 & 3.05 & - \\
        9  & Qwen2.5-Math-1.5B-Instruct & 51.80 & 51.30 & 0.51 & - \\
        10 & Qwen2.5-Math-1.5B-OREO & 39.92 & 38.45 & 1.47 & - \\
        \bottomrule
    \end{tabular}
    \caption{\footnotesize Model rankings and score comparison between MATH-500 and MATH-100. The score difference was computed as the absolute difference between the MATH-500 and MATH-100 scores. The rank difference indicates the change in ranking on MATH-100 relative to the performance on MATH-500.}
    \label{tab:sampled_math}
\end{table*}

In Table~\ref{tab:sampled_math}, we report the performance of the evaluated models. The score differences are relatively small, and even after accounting for minor variations, the ranking of the 10 models remains largely consistent—with only a few instances of rank switching. We conclude that the sampled version serves as an acceptable proxy for the full dataset and proceed accordingly.

\subsection{Sourcing M-IMO and M-MO}\label{app:mimo_mmo}

\begin{figure}[h]
\centering
\includegraphics[width=\columnwidth]{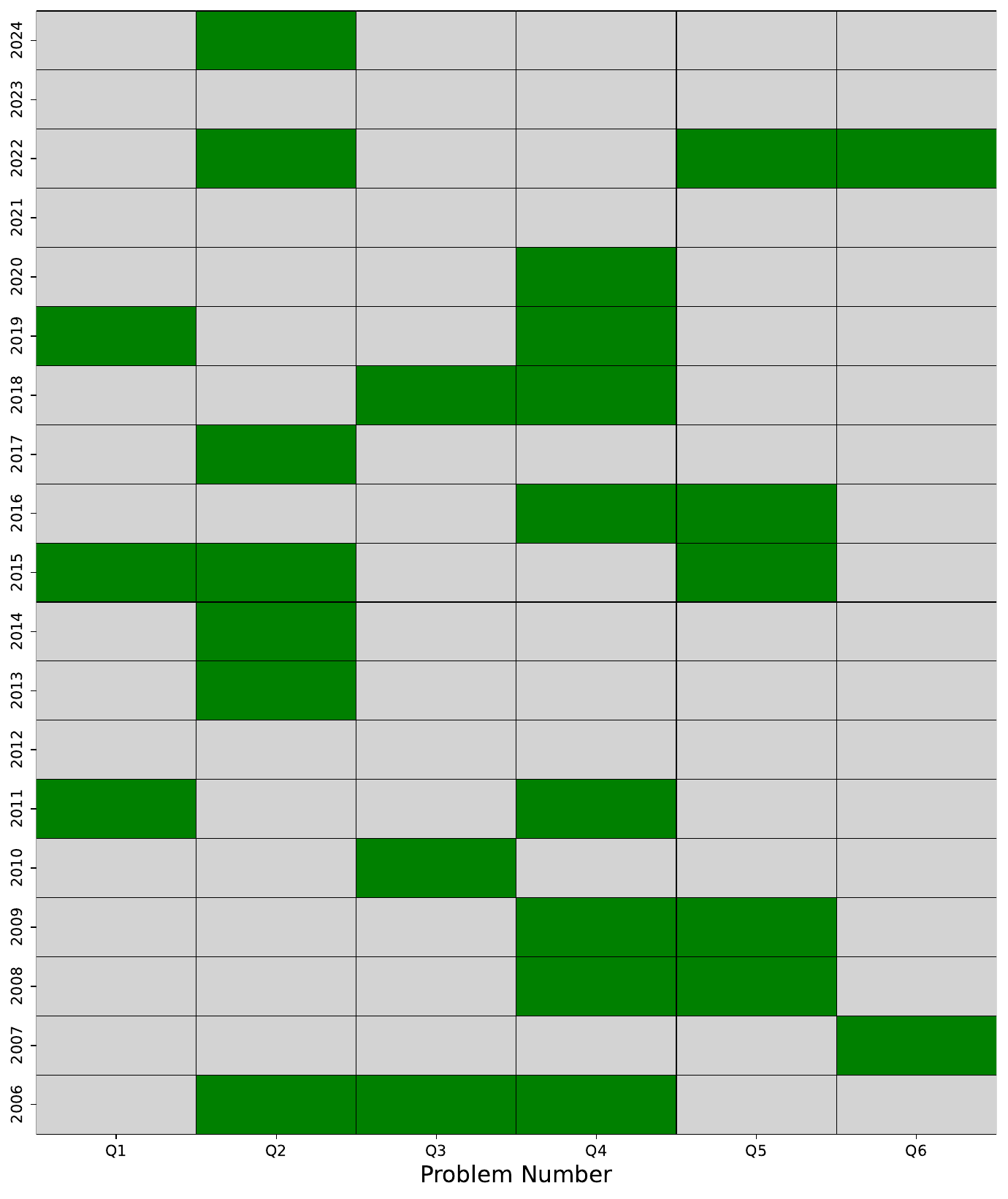}
\caption{\footnotesize  Heatmap representation of IMO problems from 2006 to 2024. Each row corresponds to a competition year, and each column represents a problem (Q1–Q6). Green cells indicate questions that have been included in the M-IMO subset, while gray cells represent problems that were not selected.}
\label{fig_imo}
\end{figure}

Relying solely on machine-translated benchmarks (MT-MATH100 and MT-AIME2024) carries inherent risks. To mitigate this, we supplement our dataset with questions from the International Mathematical Olympiad (IMO) and various regional math olympiads. Figure~\ref{fig_imo} provides an overview of the IMO questions included from 2004 to 2024, while Table~\ref{tab:math_competitions} lists the sources for regional olympiads. For English, Chinese, and Korean, we utilize existing datasets rather than recollecting questions~\citep{he2024olympiadbench, ko2025understand}.

\begin{table*}[h]
\centering
\begin{tabular}{l|p{10cm}}
\toprule
\textbf{Language} & \textbf{Competition Links} \\
\midrule
French & \url{https://euler.ac-versailles.fr/spip.php?rubrique207} \\
German & DeMO \\
Japanese & \url{https://www.imojp.org/domestic/jmo_overview.html#Problems} \\
Dutch & \url{https://prime.ugent.be/activiteiten/puma/} \newline \url{https://wiskundeolympiade.nl/wedstrijdarchief/1e-ronde} \\
Czech & \url{https://www.matematickaolympiada.cz/mo-pro-ss/rocnik} \newline \url{https://iksko.org/problems.php} \\
Polish & \url{https://om.sem.edu.pl/problems/} \\
Slovakian & \url{https://skmo.sk/dokumenty.php?rocnik=74} \newline \url{https://riesky.sk/archiv/} \\
Russian & \url{https://mmo.mccme.ru//} \\
\bottomrule
\end{tabular}
\caption{Link to mathematical competition links that has been included in M-MO subset.}
\label{tab:math_competitions}
\end{table*}

\subsection{Prompts}\label{app:prompts}

The prompts used for question translation (Figure~\ref{question_system}), solution translation (Figure~\ref{solution_template}), and model judgment (Figure~\ref{judge_template}) are detailed accordingly.

\subsection{Benchmark Contamination}\label{app:contamination}
As LLMs are trained on ever-growing datasets, concerns about benchmark contamination have emerged~\citep{xu2024benchmark}. Because most training data and procedures remain proprietary, it is virtually impossible to guarantee that a benchmark is entirely free of contamination. Existing detection methods remain unstable, and our \textsc{MCLM} dataset, particularly its \textsc{M-IMO} and \textsc{M-MO} subsets collected from the Internet, may be prone to exposure in various model-training corpora. For instance, we observe that \emph{Llama-3.1-3B-Instruct} can produce correct answers without any visible reasoning, suggesting prior familiarity with certain questions. However, lacking a robust decontamination method, we have chosen not to remove potentially contaminated samples. Despite this possibility, the overall low performance of most models on these subsets suggests that \textsc{MCLM} remains a challenging and valuable resource for evaluating multilingual, competition-level math. Furthermore, our primary focus is on assessing \emph{multilingual consistency}. If a model correctly ``guesses'' a single language version of a question (possibly due to prior exposure), it does not necessarily indicate strong multilingual reasoning. Conversely, if the model can solve all language variations after seeing only a few, it may demonstrate a degree of cross-lingual robustness.

\subsection{License}
The machine-translated subsets of MCLM are released under the MIT License. The remaining subsets are provided under a CC BY-NC-ND license, although we may transition to a more permissive license pending further review of the original competition data policies.

\section{Additional details in training LLMs with system 2 thinking}
\label{app_training}
In this section, we provide additional details on the dataset used to train self-correcting LLMs (Section~\ref{app_comp_w_past}), ablation studies on MR1 training (Section~\ref{app_ablation_training}), and the training configurations.

\subsection{Additional details on the training dataset}~\label{app_comp_w_past}
In training MR1 (i.e., Deepseek-R1-1.5B + SFT with translated data), we first collect thinking trajectories generated by \textsc{R1} \citep{guo2025deepseek} from the Numina Math dataset in \emph{Be-Spoke Stratos} \citep{bespokestratos} and \emph{OpenThoughts} \citep{openthoughts}. To ensure high-quality reasoning supervision, we exclude any data distilled from smaller thinking-model variants (e.g., \emph{QwQ-32B-Preview} \citep{qwq-32b-preview} or the \emph{R1-Distil} series). We then employ GPT-4o as an LLM-based judge to filter out instances with incorrect answers. 

Next, the question and solution for each remaining instance are translated into one of the 14 languages in Group (B). We opt for only 14 languages---as opposed to all 55---because our ablation studies indicate that oversampling languages can negatively impact overall performance. A rule-based parser validates that the question and answer content remains unchanged post-translation. Beyond this verification, no additional quality checks are performed. However, when we encounter elevated loss spikes during training, we backtrack through the dataset to identify and remove problematic instances. Following this process, we retain approximately 120K instances.

\paragraph{Proxying problem difficulty}~Table~\ref{tab:dataset_comparison} compares \traindataset{} with existing multilingual math datasets: MGSM8KInstruct~\citep{chen2023breaking} and mCoT-MATH~\citep{lai2024mcot}. MGSM8KInstruct extends GSM8K~\citep{lightman2023let} by translating it into 10 languages, yielding a parallel dataset of approximately 8,000 questions per language. In contrast, mCoT-MATH sources 560,000 seed questions from MetaMathQA and MathInstruct and translates them into 10 languages.  

\begin{table}[h]
    \centering
    \fontsize{9}{11}\selectfont
    \begin{tabular}{lccc}
    \toprule
        \textbf{Dataset} & \textbf{\# Lang.} & \textbf{\# Inst.} & \textbf{Diff.} \\ 
        \midrule
        MGSM8KInstruct & 10 & 73.6k & G.S \\
        mCoT-MATH      & 10 & 6.3M  & G.S \\
        \midrule
        Euler-Instruct (Ours) & 55 & 250K & C.L \\ 
        \bottomrule
    \end{tabular}
    \caption{\footnotesize \textbf{Comparison of Multilingual Mathematical Reasoning Datasets.} The \textit{Diff.} column indicates difficulty level, where \textit{G.S} represents grade school level and \textit{C.L} represents competition level.}
    \label{tab:dataset_comparison}
\end{table}

To estimate the difficulty level of each dataset, we randomly sample 1,000 questions and measure the solve rate of LLMs. Since MGSM8KInstruct and mCoT-MATH were published before \traindataset, they may have been included in the training of existing LLMs, potentially making a direct comparison unfair. To address this, we first evaluate using OLMo2-\{7, 13\}B~\citep{olmo20242}; we use the base model instead of the instruct model since GSM8K is used during the instruction tuning phase. Since OLMo2-7B-base is not instruction-tuned, we provide three-shot examples and prompt it to solve the questions in a structured format: "The answer is X."  Additionally, we evaluate with Qwen2.5-\{7, 32, 72\}B-Instruct~\citep{yang2024qwen2} in a zero-shot setting, using a parser to extract answers.

\begin{figure}[h]
\centering
\includegraphics[width=\columnwidth]{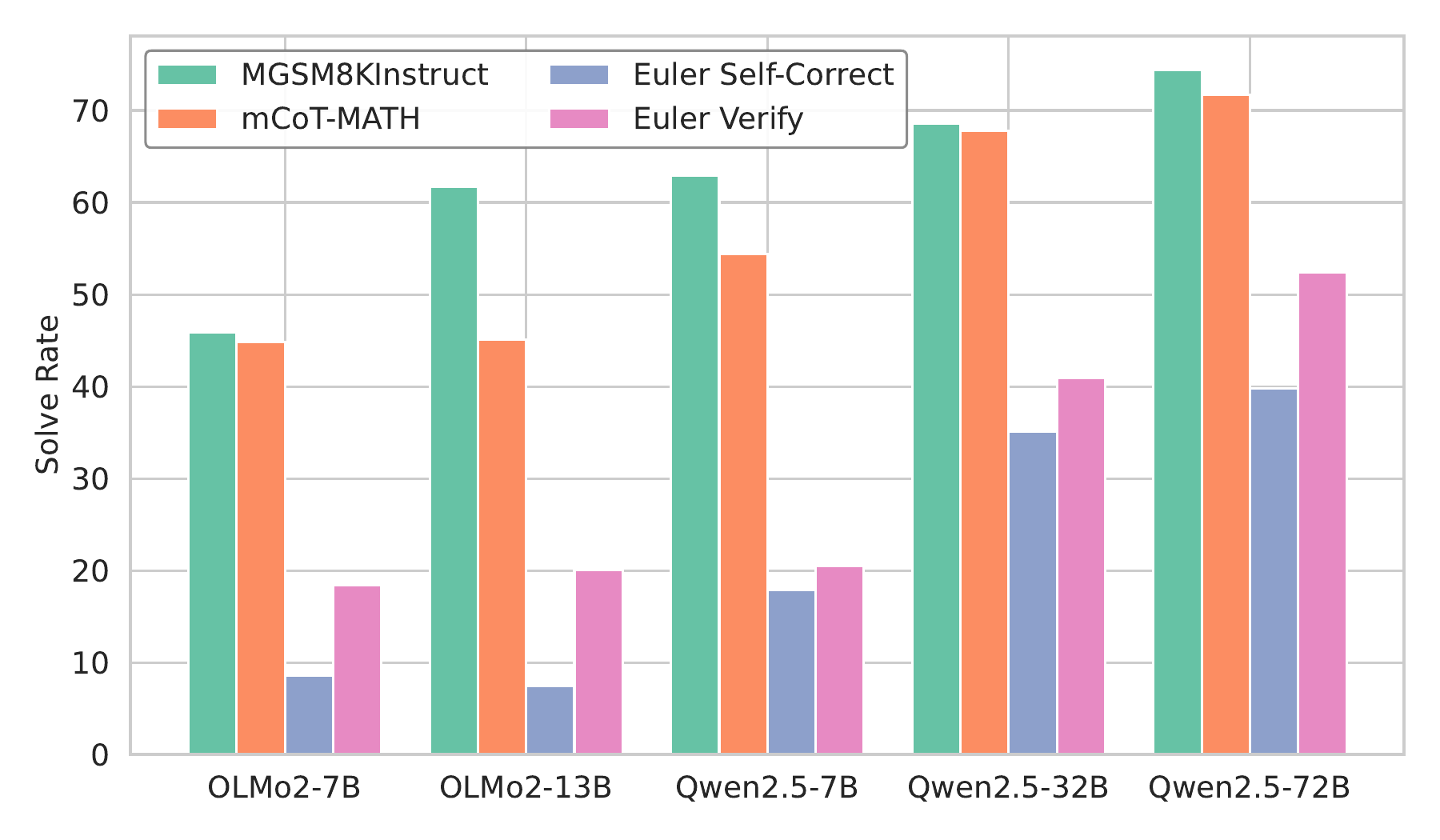}
\caption{\footnotesize \textbf{Solve rates (\%) of different multilingual math datasets evaluated.} For the OLMo2 series, we use the base models, while for the Qwen2.5 series, the instruct-tuned variants are used. \traindataset presents a significantly lower solve rate, indicating its greater difficulty.}
\label{fig_solve_rate}
\end{figure}

Figure~\ref{fig_solve_rate} illustrates that, for both mCoT-MATH and MGSM8KInstruct, over half of the problems are solved by Qwen2.5-7B-Instruct, and Qwen2.5-72B-Instruct solves more than 70\%. In contrast, the solve rate for \traindataset remains low across all models, with the 7B variants solving less than 20\% of the problems and even the 72B variants achieving below 40\%. Notably, the Verify subset exhibits a lower solve rate than previous datasets, indicating that its difficulty remains high even when restricted to numerical answers.

\subsection{Training Ablations}~\label{app_ablation_training}
Before training MR1 we meet the question: \emph{How little additional data is required to incorporate a new language into a self-correcting model?} To address this, we train four models under a fixed total budget of 24,000 training instances. Table~\ref{tab:language_distribution} details the language composition and the per-language instance allocation for each model.
\begin{table}[ht]
\centering
\fontsize{9}{11}\selectfont
\begin{tabular}{l c c}
\toprule
\textbf{Languages} & \textbf{\# Lang.} & \textbf{\# Instances} \\
\midrule
{\small ko} & 1 & 24k \\
{\small af, fr, ko} & 3 & 8k \\
{\small af, ar, fr, he, id, ko, tr} & 7 & $\approx$3.5k \\
{\small all 14 in \traindataset} & 14 & $\approx$1.7k \\
\bottomrule
\end{tabular}
\caption{\footnotesize \textbf{Details on trained models.} All models are trained with a total of 24,000 instances. \# Instances denote the number of instances used per language.}
\label{tab:language_distribution}
\end{table}

\begin{figure*}[t!]
\centering
\includegraphics[width=\textwidth]{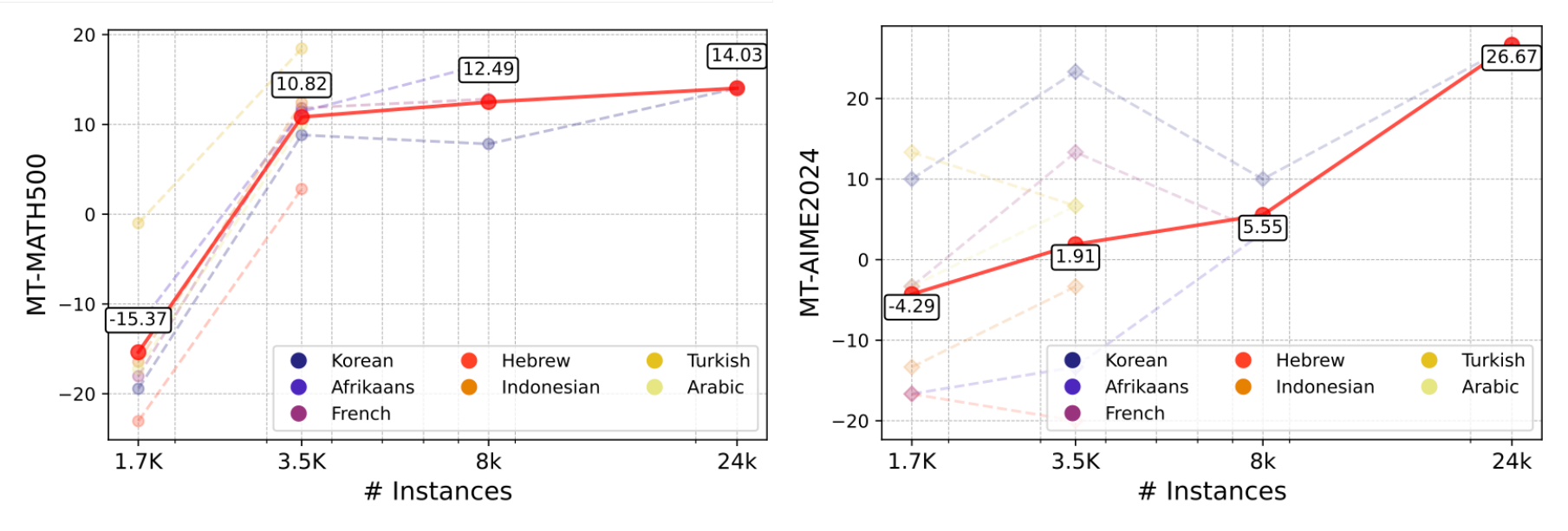}
\caption{\footnotesize \textbf{Model Results from Table~\ref{tab:language_distribution}.} Left shows accuracy on MT-MATH500 (entire translated subset for language group (B)), and right shows average performance of MT-AIME2024.}
\label{fig_ablation}
\end{figure*}

Figure~\ref{fig_ablation} presents the performance gains over the base model (DeepSeek-R1-1.5B) as a function of the number of training instances. On MT-MATH100, performance rises sharply once the per-language budget reaches approximately 3.5k instances. By contrast, MT-AIME2024 shows more gradual improvements, suggesting that transferring self-correction capabilities to a challenging set of new-language questions requires a larger data allocation. Based on these findings, we use 14 languages within a total budget of 120K, ensuring that each language includes at least 8k instances.

\subsection{Training Configurations and Logs}

Axolotl~\citep{axolotl2025} is used for the SFT training in Section~\ref{budget_forcing}. We train Qwen2.5-Math-1.5B with DeepSpeed-Zero1~\citep{rajbhandari2020zero} on 4 A100 80GB GPUs for 8 hours per run. \citet{hsu2024liger} is used for optimization.
\begin{table}[hb]
\centering
\fontsize{9}{11} \selectfont
\begin{tabular}{cc}
\toprule
\textbf{Category} & \textbf{Section~\ref{budget_forcing}} \\ \midrule
\textbf{Sequence Length} & 16,384 \\
\textbf{Learning Rate} & \(2 \times 10^{-5}\) \\
\textbf{Global Batch (Effective)} & 128 \\
\textbf{Learning Rate Scheduler} & Cosine Decay \\
\textbf{Warmup Ratio} & 0.05 \\
\textbf{Training Epochs} & 3 \\ \bottomrule
\end{tabular}
\caption{\footnotesize \textbf{SFT configuration details for Section~\ref{budget_forcing}.}}
\label{tab:training_configs}
\end{table}

\begin{figure*}[h]
\begin{instructionsbox}
\begin{lstlisting}
You will be given an English question in the following format.

[Question]
<question>
{..question...}
</question>

Your job is to return a translated version of the question.
* Translate to <language>.
* The translation must be fluent, easy to read by native speakers.
* Do not solve the prompt translate it.
* You must preserve all details including math notations (latex) and code. 
* The math notations and code must not be translated, keep it as is.
* Return you translation in the following format.

[Translation]
<translation>
{..translated question...}
</translation>

--------------------------------------------------
The following is the math problem for you task:

[Question]
<question>
<source_question>
</question>
\end{lstlisting}
\end{instructionsbox}
\caption{Question Translation Template}
\label{question_system}
\end{figure*}

\begin{figure*}[h]
\begin{instructionsbox}
\begin{lstlisting}
You will be given an English solution in the following format.

[Solution]
<solution>
{..solution in English...}
</solution>

Your job is to rewrite the English solution to <language>. 
* The solution must preserve the original structure and details.
* You must preserve all details including math notations (latex) and code. 
* The math notations and code must not be translated, keep it as is.
* The solution must be natural, easy and polite for a native speaker to read.

[Translation]
<translation>
{..translated solution...}
</translation>

--------------------------------------------------
The following is the math problem and solution for your task:

[Solution]
<solution>
<source_solution>
</solution>
\end{lstlisting}
\end{instructionsbox}
\caption{Solution Translation Template}
\label{solution_template}
\end{figure*}

\begin{figure*}[h]
\begin{instructionsbox}
\begin{lstlisting}
You will be given a math problem, the correct answer, and a solution generated by a language model. Your task is to determine whether the solution generated by the model is correct.

[Question]
<question>
{..math question...}
</question>

[Correct Answer]
<answer>
{..correct answer...}
</answer>

[Model Solution]
<solution>
{..model-generated solution...}
</solution>

Instructions:
* Compare the model's solution with the correct answer.
* If the model's solution is correct, output [[TRUE]].
* If the model's solution is incorrect, output [[FALSE]].
* You do not have to judge the solution process; there are numerous possible 'Gold' solutions, and the model solution does not have to be identical with the one provided. As long as the model reaches the correct answer, it is correct.
* Do not provide any explanations -- only return your judgment ONLY.

--------------------------------------------------
The following is the math problem and solution for your task:

[Question]  
<question>
<math_question>
</question>

[Correct Answer]
<answer>
<correct_answer>
</answer>

[Model Solution]
<solution>
<model_solution>
</solution>
\end{lstlisting}
\end{instructionsbox}
\caption{Judge Template}
\label{judge_template}
\end{figure*}

\section{Additional Results}\label{app_add_results}
From Tables \ref{tab:1_5B_greedy} to \ref{tab:o3_mini_result}, we report detailed evaluation results for 55 languages with varying models and test-time scaling methods.
\begin{table*}[]
\centering
\fontsize{9}{11} \selectfont
\begin{tabular}{c|cccl}
\toprule
\textbf{Language} & \textbf{MT-MATH100} & \textbf{MT-AIME2024} & \textbf{M-IMO} & \multicolumn{1}{c}{\textbf{M-MO}} \\ \midrule
Afrikaans & 47.47 & 20.00 & 11.11 &  \\
Albanian & 45.45 & 10.00 & 4.00 &  \\
Arabic & 38.38 & 30.00 & 11.11 &  \\
Bengali & 37.37 & 3.33 & \multicolumn{1}{l}{} &  \\
Bulgarian & 39.39 & 13.33 & 7.41 &  \\
Catalan & 50.51 & 23.33 & \multicolumn{1}{l}{} &  \\
Chinese (Simplified) & 63.64 & 26.67 & 18.52 & \multicolumn{1}{c}{40.00} \\
Chinese (Traditional) & 61.62 & 20.00 & 18.52 &  \\
Croatian & 49.49 & 20.00 & 7.41 &  \\
Czech & 44.44 & 13.33 & 14.81 & \multicolumn{1}{c}{6.67} \\
Danish & 53.54 & 16.67 & 22.22 &  \\
Dutch & 50.51 & 36.67 & 11.11 & \multicolumn{1}{c}{20.00} \\
Estonian & 39.39 & 10.00 & 4.00 &  \\
Finnish & 41.41 & 16.67 & 8.00 &  \\
French & 62.63 & 30.00 & 18.52 & \multicolumn{1}{c}{51.61} \\
German & 47.47 & 26.67 & 11.11 & \multicolumn{1}{c}{10.00} \\
Greek & 33.33 & 13.33 & 5.26 &  \\
Gujarati & 39.39 & 10.00 & \multicolumn{1}{l}{} &  \\
Hebrew & 38.38 & 13.33 & 3.70 &  \\
Hindi & 35.35 & 6.67 & \multicolumn{1}{l}{} &  \\
Hungarian & 51.52 & 10.00 & 8.00 &  \\
Indonesian & 56.57 & 16.67 & 14.29 &  \\
Italian & 51.52 & 20.00 & 20.00 &  \\
Japanese & 56.57 & 16.67 & 8.00 & \multicolumn{1}{c}{0.00} \\
Kannada & 37.37 & 10.00 & \multicolumn{1}{l}{} &  \\
Korean & 44.44 & 13.33 & 3.70 & \multicolumn{1}{c}{36.67} \\
Latvian & 40.40 & 10.00 & 12.00 &  \\
Lithuanian & 45.45 & 6.67 & 18.52 &  \\
Macedonian & 43.43 & 10.00 & 11.11 &  \\
Malayalam & 43.43 & 23.33 & \multicolumn{1}{l}{} &  \\
Marathi & 34.34 & 13.33 & \multicolumn{1}{l}{} &  \\
Nepali & 36.36 & 6.67 & \multicolumn{1}{l}{} &  \\
Norwegian & 53.54 & 23.33 & 11.11 &  \\
Persian & 38.38 & 10.00 & \multicolumn{1}{l}{} &  \\
Polish & 54.55 & 26.67 & 14.81 & \multicolumn{1}{c}{26.67} \\
Portuguese & 55.56 & 10.00 & 24.00 &  \\
Punjabi & 37.37 & 16.67 & \multicolumn{1}{l}{} &  \\
Romanian & 49.49 & 13.33 & 25.93 &  \\
Russian & 59.60 & 20.00 & 16.00 & \multicolumn{1}{c}{20.00} \\
Slovak & 48.48 & 20.00 & 11.11 & \multicolumn{1}{c}{6.67} \\
Slovenian & 49.49 & 10.00 & 14.81 &  \\
Somali & 42.42 & 23.33 & \multicolumn{1}{l}{} &  \\
Spanish & 55.56 & 20.00 & 18.52 &  \\
Swahili & 34.34 & 16.67 & \multicolumn{1}{l}{} &  \\
Swedish & 58.59 & 20.00 & 8.00 &  \\
Tagalog & 46.46 & 16.67 & \multicolumn{1}{l}{} &  \\
Tamil & 38.38 & 10.00 & \multicolumn{1}{l}{} &  \\
Telugu & 39.39 & 6.67 & \multicolumn{1}{l}{} &  \\
Thai & 39.39 & 23.33 & 3.70 &  \\
Turkish & 43.43 & 13.33 & 7.41 &  \\
Ukrainian & 38.38 & 13.33 & 11.11 &  \\
Urdu & 35.35 & 20.00 & \multicolumn{1}{l}{} &  \\
Vietnamese & 44.44 & 13.33 & 7.41 &  \\
Welsh & 39.39 & 16.67 & \multicolumn{1}{l}{} &  \\
\rowcolor[HTML]{FCE5CD}English & 67.68 & 20.00 & 18.52 & \multicolumn{1}{c}{56.67} \\ \midrule
Average & 46.01 & 16.36 & 12.23 & \multicolumn{1}{c}{25.00} \\
Standard Deviation & 8.61 & 6.89 & 6.02 & \multicolumn{1}{c}{19.10} \\
Fleiss' Kappa & 0.56 & 0.68 & 0.24 &  \\ \bottomrule
\end{tabular}
\caption{\footnotesize Evaluation results of Qwen2.5-Math-1.5B-Instruct with greedy decoding on MCLM.}
\label{tab:1_5B_greedy}
\end{table*}

\begin{table*}[]
\centering
\fontsize{7}{10} \selectfont
\begin{tabular}{c|cc|cc|cc}
\toprule
\multicolumn{1}{c|}{} & \multicolumn{2}{c|}{\textbf{ORM (K=2)}} & \multicolumn{2}{c|}{\textbf{ORM (K=4)}} & \multicolumn{2}{c}{\textbf{ORM (K=8)}} \\ \cmidrule{2-7}
\multicolumn{1}{c|}{\multirow{-2}{*}{\textbf{Language}}} & \textbf{MT-MATH100} & \multicolumn{1}{c|}{\textbf{MT-AIME2024}} & \textbf{MT-MATH100} & \multicolumn{1}{c|}{\textbf{MT-AIME2024}} & \textbf{MT-MATH100} & \textbf{MT-AIME2024} \\ \midrule
Afrikaans & 53.54 & 23.33 & 56.57 & 16.67 & 60.61 & 23.33 \\
Albanian & 52.53 & 10.00 & 50.51 & 10.00 & 47.47 & 13.33 \\
Arabic & 43.43 & 20.00 & 46.46 & 13.33 & 51.52 & 16.67 \\
Bengali & 41.41 & 10.00 & 40.40 & 10.00 & 41.41 & 13.33 \\
Bulgarian & 45.45 & 26.67 & 46.46 & 20.00 & 51.52 & 16.67 \\
Catalan & 59.60 & 33.33 & 63.64 & 33.33 & 61.62 & 26.67 \\
Chinese (Simplified) & 69.70 & 36.67 & 76.77 & 30.00 & 78.79 & 26.67 \\
Chinese (Traditional) & 68.69 & 13.33 & 70.71 & 20.00 & 74.75 & 26.67 \\
Croatian & 51.52 & 16.67 & 59.60 & 23.33 & 58.59 & 30.00 \\
Czech & 49.49 & 13.33 & 56.57 & 10.00 & 59.60 & 16.67 \\
Danish & 53.54 & 23.33 & 56.57 & 20.00 & 59.60 & 26.67 \\
Dutch & 51.52 & 30.00 & 57.58 & 26.67 & 63.64 & 23.33 \\
Estonian & 46.46 & 13.33 & 48.48 & 13.33 & 50.51 & 13.33 \\
Finnish & 41.41 & 13.33 & 48.48 & 20.00 & 53.54 & 20.00 \\
French & 64.65 & 40.00 & 68.69 & 33.33 & 73.74 & 30.00 \\
German & 54.55 & 23.33 & 63.64 & 23.33 & 64.65 & 30.00 \\
Greek & 39.39 & 13.33 & 44.44 & 10.00 & 47.47 & 10.00 \\
Gujarati & 44.44 & 10.00 & 43.43 & 16.67 & 47.47 & 13.33 \\
Hebrew & 44.44 & 16.67 & 46.46 & 13.33 & 49.49 & 10.00 \\
Hindi & 40.40 & 10.00 & 45.45 & 13.33 & 47.47 & 16.67 \\
Hungarian & 53.54 & 10.00 & 57.58 & 10.00 & 63.64 & 16.67 \\
Indonesian & 58.59 & 20.00 & 56.57 & 20.00 & 59.60 & 16.67 \\
Italian & 57.58 & 26.67 & 60.61 & 26.67 & 69.70 & 16.67 \\
Japanese & 59.60 & 16.67 & 66.67 & 23.33 & 70.71 & 26.67 \\
Kannada & 45.45 & 10.00 & 47.47 & 16.67 & 52.53 & 13.33 \\
Korean & 53.54 & 16.67 & 56.57 & 23.33 & 57.58 & 13.33 \\
Latvian & 45.45 & 10.00 & 51.52 & 20.00 & 54.55 & 16.67 \\
Lithuanian & 48.48 & 10.00 & 52.53 & 10.00 & 57.58 & 13.33 \\
Macedonian & 50.51 & 13.33 & 51.52 & 13.33 & 50.51 & 10.00 \\
Malayalam & 47.47 & 20.00 & 52.53 & 20.00 & 56.57 & 23.33 \\
Marathi & 39.39 & 13.33 & 43.43 & 23.33 & 43.43 & 20.00 \\
Nepali & 38.38 & 6.67 & 46.46 & 3.33 & 46.46 & 6.67 \\
Norwegian & 59.60 & 26.67 & 61.62 & 16.67 & 65.66 & 23.33 \\
Persian & 40.40 & 13.33 & 41.41 & 13.33 & 39.39 & 16.67 \\
Polish & 54.55 & 16.67 & 57.58 & 16.67 & 64.65 & 16.67 \\
Portuguese & 58.59 & 13.33 & 60.61 & 13.33 & 62.63 & 26.67 \\
Punjabi & 41.41 & 16.67 & 43.43 & 20.00 & 42.42 & 16.67 \\
Romanian & 51.52 & 23.33 & 54.55 & 23.33 & 56.57 & 20.00 \\
Russian & 60.61 & 20.00 & 65.66 & 23.33 & 68.69 & 23.33 \\
Slovak & 52.53 & 10.00 & 54.55 & 20.00 & 55.56 & 33.33 \\
Slovenian & 47.47 & 16.67 & 51.52 & 20.00 & 54.55 & 30.00 \\
Somali & 44.44 & 16.67 & 46.46 & 16.67 & 46.46 & 10.00 \\
Spanish & 58.59 & 23.33 & 65.66 & 26.67 & 68.69 & 30.00 \\
Swahili & 37.37 & 13.33 & 41.41 & 20.00 & 45.45 & 13.33 \\
Swedish & 57.58 & 20.00 & 59.60 & 23.33 & 60.61 & 20.00 \\
Tagalog & 50.51 & 16.67 & 55.56 & 20.00 & 57.58 & 23.33 \\
Tamil & 41.41 & 16.67 & 44.44 & 16.67 & 47.47 & 16.67 \\
Telugu & 42.42 & 13.33 & 46.46 & 20.00 & 48.48 & 20.00 \\
Thai & 44.44 & 10.00 & 49.49 & 20.00 & 57.58 & 13.33 \\
Turkish & 50.51 & 16.67 & 46.46 & 13.33 & 54.55 & 20.00 \\
Ukrainian & 44.44 & 23.33 & 51.52 & 16.67 & 52.53 & 26.67 \\
Urdu & 38.38 & 16.67 & 41.41 & 16.67 & 44.44 & 20.00 \\
Vietnamese & 49.49 & 23.33 & 50.51 & 30.00 & 52.53 & 33.33 \\
Welsh & 38.38 & 16.67 & 44.44 & 16.67 & 44.44 & 20.00 \\
\rowcolor[HTML]{FCE5CD} 
English & 71.72 & 16.67 & 73.74 & 26.67 & 76.77 & 36.67 \\ \midrule
Average & 50.01 & 17.64 & 53.50 & 18.85 & 56.25 & 20.12 \\
Standard Deviation & 8.47 & 7.05 & 8.83 & 6.23 & 9.50 & 6.97 \\
Fleiss' Kappa & 0.57 & 0.66 & 0.60 & 0.64 & 0.61 & 0.63 \\ \bottomrule
\end{tabular}
\caption{\footnotesize Evaluation results of Qwen2.5-Math-1.5B-Instruct with Best-of-N \((K=2, 4, 8)\) using Qwen2.5-Math-RM-72B as ORM on MT-MATH100 and MT-AIME2024.}
\label{tab:1_5B_orm}
\end{table*}

\begin{table*}[]
\centering
\fontsize{6}{9} \selectfont
\begin{tabular}{c|cc|cc|cccl}
\toprule
 & \multicolumn{2}{c|}{\textbf{PRM (S=3, c=3)}} & \multicolumn{2}{c|}{\textbf{PRM (S=4, c=5)}} & \multicolumn{4}{c}{\textbf{PRM (S=5, c=8)}} \\ \cmidrule{2-9}
\multirow{-2}{*}{\textbf{Language}} & \textbf{MT-MATH100} & \textbf{MT-AIME2024} & \textbf{MT-MATH100} & \textbf{MT-AIME2024} & \textbf{MT-MATH100} & \textbf{MT-AIME2024} & \textbf{M-IMO} & \multicolumn{1}{c}{\textbf{M-MO}} \\ \midrule
Afrikaans & 52.53 & 6.67 & 57.58 & 20.00 & 64.65 & 10.00 & 22.73 &  \\
Albanian & 44.44 & 13.33 & 52.53 & 10.00 & 45.45 & 16.67 & 11.54 &  \\
Arabic & 41.41 & 13.33 & 52.53 & 13.33 & 45.45 & 10.00 & 7.41 &  \\
Bengali & 40.40 & 13.33 & 44.44 & 13.33 & 41.41 & 16.67 & \multicolumn{1}{l}{} &  \\
Bulgarian & 42.42 & 20.00 & 42.42 & 10.00 & 55.56 & 10.00 & 11.11 &  \\
Catalan & 55.56 & 10.00 & 66.67 & 26.67 & 61.62 & 26.67 & \multicolumn{1}{l}{} &  \\
Chinese (Simplified) & 64.65 & 13.33 & 75.76 & 16.67 & 71.72 & 33.33 & 25.93 & \multicolumn{1}{c}{} \\
Chinese (Traditional) & 63.64 & 26.67 & 73.74 & 16.67 & 72.73 & 26.67 & 29.63 & \multicolumn{1}{c}{\multirow{-2}{*}{53.33}} \\
Croatian & 50.51 & 13.33 & 51.52 & 20.00 & 54.55 & 23.33 & 14.81 &  \\
Czech & 50.51 & 10.00 & 52.53 & 16.67 & 58.59 & 20.00 & 14.81 & \multicolumn{1}{c}{10.00} \\
Danish & 57.58 & 10.00 & 60.61 & 30.00 & 60.61 & 20.00 & 22.22 &  \\
Dutch & 56.57 & 20.00 & 56.57 & 26.67 & 59.60 & 20.00 & 7.41 & \multicolumn{1}{c}{20.00} \\
Estonian & 47.47 & 13.33 & 51.52 & 3.33 & 49.49 & 10.00 & 11.54 &  \\
Finnish & 41.41 & 10.00 & 43.43 & 6.67 & 49.49 & 10.00 & 15.38 &  \\
French & 62.63 & 13.33 & 65.66 & 30.00 & 70.71 & 20.00 & 18.52 & \multicolumn{1}{c}{51.61} \\
German & 54.55 & 40.00 & 62.63 & 30.00 & 58.59 & 23.33 & 22.22 & \multicolumn{1}{c}{16.67} \\
Greek & 42.42 & 13.33 & 39.39 & 6.67 & 44.44 & 20.00 & 4.35 &  \\
Gujarati & 42.42 & 6.67 & 39.39 & 13.33 & 41.41 & 13.33 & \multicolumn{1}{l}{} &  \\
Hebrew & 46.46 & 6.67 & 42.42 & 23.33 & 47.47 & 6.67 & 7.41 &  \\
Hindi & 39.39 & 10.00 & 46.46 & 20.00 & 47.47 & 10.00 & \multicolumn{1}{l}{} &  \\
Hungarian & 57.58 & 26.67 & 61.62 & 10.00 & 57.58 & 3.33 & 19.23 &  \\
Indonesian & 56.57 & 16.67 & 57.58 & 13.33 & 64.65 & 13.33 & 20.83 &  \\
Italian & 61.62 & 13.33 & 61.62 & 20.00 & 67.68 & 23.33 & 23.08 &  \\
Japanese & 64.65 & 20.00 & 66.67 & 26.67 & 66.67 & 16.67 & 15.38 & \multicolumn{1}{c}{7.14} \\
Kannada & 44.44 & 23.33 & 42.42 & 13.33 & 47.47 & 13.33 & \multicolumn{1}{l}{} &  \\
Korean & 46.46 & 10.00 & 45.45 & 13.33 & 50.51 & 13.33 & 14.81 & \multicolumn{1}{c}{26.67} \\
Latvian & 47.47 & 6.67 & 50.51 & 16.67 & 51.52 & 10.00 & 15.38 &  \\
Lithuanian & 42.42 & 10.00 & 49.49 & 6.67 & 45.45 & 16.67 & 14.81 &  \\
Macedonian & 41.41 & 13.33 & 47.47 & 16.67 & 48.48 & 23.33 & 11.11 &  \\
Malayalam & 38.38 & 16.67 & 42.42 & 16.67 & 43.43 & 13.33 & \multicolumn{1}{l}{} &  \\
Marathi & 39.39 & 10.00 & 43.43 & 10.00 & 36.36 & 13.33 & \multicolumn{1}{l}{} &  \\
Nepali & 41.41 & 16.67 & 41.41 & 26.67 & 42.42 & 10.00 & \multicolumn{1}{l}{} &  \\
Norwegian & 59.60 & 23.33 & 65.66 & 30.00 & 59.60 & 26.67 & 18.52 &  \\
Persian & 37.37 & 20.00 & 43.43 & 13.33 & 39.39 & 13.33 & \multicolumn{1}{l}{} &  \\
Polish & 49.49 & 23.33 & 58.59 & 23.33 & 62.63 & 20.00 & 25.93 & \multicolumn{1}{c}{36.67} \\
Portuguese & 58.59 & 20.00 & 57.58 & 16.67 & 61.62 & 30.00 & 19.23 &  \\
Punjabi & 39.39 & 20.00 & 40.40 & 13.33 & 49.49 & 6.67 & \multicolumn{1}{l}{} &  \\
Romanian & 57.58 & 16.67 & 55.56 & 13.33 & 57.58 & 10.00 & 22.22 &  \\
Russian & 53.54 & 23.33 & 65.66 & 23.33 & 64.65 & 20.00 & 15.38 & \multicolumn{1}{c}{23.33} \\
Slovak & 51.52 & 10.00 & 52.53 & 13.33 & 53.54 & 20.00 & 14.81 &  \\
Slovenian & 44.44 & 23.33 & 47.47 & 16.67 & 45.45 & 26.67 & 11.11 &  \\
Somali & 43.43 & 6.67 & 42.42 & 23.33 & 40.40 & 3.33 & \multicolumn{1}{l}{} &  \\
Spanish & 60.61 & 16.67 & 65.66 & 26.67 & 72.73 & 30.00 & 29.63 &  \\
Swahili & 38.38 & 13.33 & 41.41 & 13.33 & 41.41 & 10.00 & \multicolumn{1}{l}{} &  \\
Swedish & 55.56 & 13.33 & 57.58 & 13.33 & 57.58 & 20.00 & 15.38 &  \\
Tagalog & 47.47 & 20.00 & 51.52 & 10.00 & 55.56 & 10.00 & \multicolumn{1}{l}{} &  \\
Tamil & 41.41 & 10.00 & 45.45 & 16.67 & 45.45 & 16.67 & \multicolumn{1}{l}{} &  \\
Telugu & 42.42 & 6.67 & 45.45 & 13.33 & 48.48 & 16.67 & \multicolumn{1}{l}{} &  \\
Thai & 39.39 & 6.67 & 47.47 & 6.67 & 50.51 & 10.00 & 14.81 &  \\
Turkish & 45.45 & 13.33 & 50.51 & 23.33 & 45.45 & 10.00 & 11.11 &  \\
Ukrainian & 39.39 & 6.67 & 45.45 & 23.33 & 51.52 & 6.67 & 18.52 &  \\
Urdu & 39.39 & 20.00 & 40.40 & 16.67 & 42.42 & 13.33 & \multicolumn{1}{l}{} &  \\
Vietnamese & 47.47 & 26.67 & 53.54 & 20.00 & 51.52 & 13.33 & 29.63 &  \\
Welsh & 43.43 & 10.00 & 48.48 & 6.67 & 51.52 & 6.67 & \multicolumn{1}{l}{} &  \\
\rowcolor[HTML]{FCE5CD}English & 73.74 & 26.67 & 79.80 & 23.33 & 72.73 & 23.33 & 29.63 & \multicolumn{1}{c}{60.00} \\ \midrule
Average & 48.87 & 15.33 & 52.54 & 17.15 & 53.54 & 16.00 & 17.31 & \multicolumn{1}{c}{30.54} \\
Standard Deviation & 8.76 & 6.93 & 9.98 & 6.95 & 9.71 & 7.15 & 6.44 & \multicolumn{1}{c}{18.88} \\
Fleiss' Kappa & 0.57 & 0.78 & 0.58 & 0.61 & 0.60 & 0.62 & 0.43 & \\ \bottomrule
\end{tabular}
\caption{\footnotesize Evaluation results of Qwen2.5-Math-1.5B-Instruct using Qwen2.5-Math-PRM-72B as PRM on MCLM.}
\label{tab:1_5B_prm_72B}
\end{table*}

\begin{table*}[]
\centering
\fontsize{10}{13} \selectfont
\begin{tabular}{c|c|c|c}
\toprule
 & \multicolumn{3}{c}{\textbf{MT-MATH100}} \\ \cmidrule{2-4}
\multirow{-2}{*}{\textbf{Language}} & \multicolumn{1}{c|}{\textbf{PRM (S=7, c=5)}} & \multicolumn{1}{c|}{\textbf{PRM (S=7, c=7)}} & \textbf{PRM (S=7, c=11)} \\ \midrule
Afrikaans & 55.56 & 51.52 & 58.59 \\
Arabic & 44.44 & 42.42 & 44.44 \\
Chinese (Simplified) & 71.72 & 74.75 & 76.77 \\
French & 64.65 & 72.73 & 69.70 \\
German & 57.58 & 58.59 & 58.59 \\
Hebrew & 46.46 & 39.39 & 44.44 \\
Indonesian & 59.60 & 62.63 & 61.62 \\
Italian & 60.61 & 60.61 & 58.59 \\
Japanese & 67.68 & 67.68 & 63.64 \\
Korean & 48.48 & 45.45 & 50.51 \\
Spanish & 64.65 & 67.68 & 68.69 \\
Turkish & 50.51 & 53.54 & 48.48 \\
Vietnamese & 51.52 & 49.49 & 51.52 \\
\rowcolor[HTML]{FCE5CD} 
English & 75.76 & 79.80 & 74.75 \\ \midrule
Average & 58.51 & 59.02 & 59.31 \\
Standard Deviation & 9.62 & 12.57 & 10.60 \\
Fleiss' Kappa & 0.56 & 0.57 & 0.56 \\ \bottomrule
\end{tabular}
\caption{\footnotesize Evaluation results of Qwen2.5-Math-1.5B-Instruct using Qwen2.5-Math-PRM-72B as PRM with steps fixed at \((S=7)\) on MT-MATH100.}
\label{tab:1_5B_prm_72B_s7}
\end{table*}

\begin{table*}[]
\centering
\fontsize{10}{13} \selectfont
\begin{tabular}{c|c|c|c}
\toprule
 & \multicolumn{3}{c}{\textbf{MT-MATH100}} \\ \cmidrule{2-4}
\multirow{-2}{*}{\textbf{Language}} & \multicolumn{1}{c|}{\textbf{PRM (S=3, c=8)}} & \multicolumn{1}{c|}{\textbf{PRM (S=6, c=8)}} & \multicolumn{1}{c|}{\textbf{PRM (S=9, c=8)}} \\ \midrule
Afrikaans & 54.55 & 55.56 & 60.61 \\
Arabic & 41.41 & 44.44 & 52.53 \\
Chinese (Simplified) & 71.72 & 71.72 & 70.71 \\
French & 67.68 & 64.65 & 67.68 \\
German & 56.57 & 57.58 & 66.67 \\
Hebrew & 42.42 & 46.46 & 45.45 \\
Indonesian & 60.61 & 59.60 & 62.63 \\
Italian & 56.57 & 60.61 & 61.62 \\
Japanese & 63.64 & 67.68 & 62.63 \\
Korean & 47.47 & 48.48 & 48.48 \\
Spanish & 65.66 & 64.65 & 72.73 \\
Turkish & 53.54 & 50.51 & 49.49 \\
Vietnamese & 57.58 & 51.52 & 57.58 \\
\rowcolor[HTML]{FCE5CD} 
English & 75.76 & 75.76 & 77.78 \\ \midrule
Average & 58.23 & 58.51 & 61.18 \\
Standard Deviation & 10.22 & 9.62 & 9.65 \\
Fleiss' Kappa & 0.56 & 0.58 & 0.58 \\ \bottomrule
\end{tabular}
\caption{\footnotesize Evaluation results of Qwen2.5-Math-1.5B-Instruct using Qwen2.5-Math-PRM-72B as PRM with the number of candidates fixed at 8, on MT-MATH100.}
\label{tab:1_5B_prm_72B_c8}
\end{table*}

\begin{table*}[]
\centering
\fontsize{10}{13} \selectfont
\begin{tabular}{c|c|c|c}
\toprule
 & \multicolumn{3}{c}{\textbf{MT-MATH100}} \\ \cmidrule{2-4}
\multirow{-2}{*}{\textbf{Language}} & \multicolumn{1}{c|}{\textbf{PRM (S=7, c=7)}} & \multicolumn{1}{c|}{\textbf{PRM (S=7, c=11)}} & \multicolumn{1}{c}{\textbf{PRM (S=7, c=18)}} \\ \midrule
Afrikaans & 51.52 & 58.59 & 58.59 \\
Arabic & 42.42 & 44.44 & 52.53 \\
Chinese (Simplified) & 74.75 & 76.77 & 76.77 \\
French & 72.73 & 69.70 & 71.72 \\
German & 58.59 & 58.59 & 60.61 \\
Hebrew & 39.39 & 44.44 & 41.41 \\
Indonesian & 62.63 & 61.62 & 62.63 \\
Italian & 60.61 & 58.59 & 64.65 \\
Japanese & 67.68 & 63.64 & 61.62 \\
Korean & 45.45 & 50.51 & 50.51 \\
Spanish & 67.68 & 68.69 & 68.69 \\
Turkish & 53.54 & 48.48 & 52.53 \\
Vietnamese & 49.49 & 51.52 & 51.52 \\
\rowcolor[HTML]{FCE5CD}English & 79.80 & 74.75 & 70.71 \\ \midrule
Average & 59.02 & 59.31 & 60.32 \\
Standard Deviation & 12.57 & 10.60 & 9.84 \\
Fleiss' Kappa & 0.52 & 0.55 & 0.54 \\ \bottomrule
\end{tabular}
\caption{\footnotesize Evaluation results of Qwen2.5-Math-1.5B-Instruct using Qwen2.5-Math-PRM-7B as PRM with the number of candidates fixed at 7, on MT-MATH100.}
\label{tab:1_5B_prm_7B_s7}
\end{table*}

\begin{table*}[]
\centering
\fontsize{10}{13} \selectfont
\begin{tabular}{c|c|c|c}
\toprule
 & \multicolumn{3}{c}{\textbf{MT-MATH100}} \\ \cmidrule{2-4}
\multirow{-2}{*}{\textbf{Language}} & \multicolumn{1}{c|}{\textbf{PRM (S=3, c=13)}} & \multicolumn{1}{c|}{\textbf{PRM (S=6, c=13)}} & \multicolumn{1}{c}{\textbf{PRM (S=9, c=13)}} \\ \midrule
Afrikaans & 55.56 & 59.60 & 54.55 \\
Arabic & 44.44 & 45.45 & 44.44 \\
Chinese (Simplified) & 75.76 & 70.71 & 79.80 \\
French & 64.65 & 71.72 & 73.74 \\
German & 55.56 & 63.64 & 61.62 \\
Hebrew & 46.46 & 43.43 & 47.47 \\
Indonesian & 56.57 & 58.59 & 61.62 \\
Italian & 62.63 & 60.61 & 61.62 \\
Japanese & 58.59 & 67.68 & 59.60 \\
Korean & 49.49 & 48.48 & 51.52 \\
Spanish & 60.61 & 73.74 & 64.65 \\
Turkish & 49.49 & 50.51 & 49.49 \\
Vietnamese & 52.53 & 48.48 & 45.45 \\
\rowcolor[HTML]{FCE5CD} 
English & 71.72 & 73.74 & 77.78 \\ \midrule
Average & 57.43 & 59.74 & 59.52 \\
Standard Deviation & 9.10 & 10.90 & 11.59 \\
Fleiss' Kappa & 0.54 & 0.55 & 0.52 \\ \bottomrule
\end{tabular}
\caption{\footnotesize Evaluation results of Qwen2.5-Math-1.5B-Instruct using Qwen2.5-Math-PRM-7B as PRM with the number of candidates fixed at 13, on MT-MATH100.}
\label{tab:1_5B_prm_7B_c13}
\end{table*}

\begin{table*}[]
\centering
\fontsize{9}{11} \selectfont
\begin{tabular}{c|cccc}
\toprule
\textbf{Language} & \textbf{MT-MATH100} & \textbf{MT-AIME2024} & \textbf{M-IMO} & \textbf{M-MO} \\ \midrule
Afrikaans & 47.47 & 36.67 & 5.56 &  \\
Albanian & 31.31 & 13.33 & 8.00 &  \\
Arabic & 36.36 & 23.33 & 7.41 &  \\
Bengali & 33.33 & 10.00 & \multicolumn{1}{l}{} &  \\
Bulgarian & 41.41 & 10.00 & 11.11 &  \\
Catalan & 47.47 & 16.67 & \multicolumn{1}{l}{} &  \\
Chinese (Simplified) & 57.58 & 23.33 & 18.52 &  \\
Chinese (Traditional) & 43.43 & 16.67 & 22.22 & \multirow{-2}{*}{23.33} \\
Croatian & 38.38 & 16.67 & 7.41 &  \\
Czech & 33.33 & 30.00 & 3.70 & 3.33 \\
Danish & 41.41 & 23.33 & 7.41 &  \\
Dutch & 45.45 & 16.67 & 7.41 & 16.67 \\
Estonian & 38.38 & 10.00 & 12.00 &  \\
Finnish & 30.30 & 23.33 & 12.00 &  \\
French & 39.39 & 6.67 & 7.41 & 35.48 \\
German & 45.45 & 23.33 & 18.52 & 6.67 \\
Greek & 30.30 & 16.67 & 0.00 &  \\
Gujarati & 27.27 & 6.67 & \multicolumn{1}{l}{} &  \\
Hebrew & 36.36 & 16.67 & 7.41 &  \\
Hindi & 36.36 & 10.00 & \multicolumn{1}{l}{} &  \\
Hungarian & 39.39 & 16.67 & 8.00 &  \\
Indonesian & 37.37 & 13.33 & 4.76 &  \\
Italian & 41.41 & 13.33 & 12.00 &  \\
Japanese & 45.45 & 20.00 & 12.00 & 3.57 \\
Kannada & 32.32 & 10.00 & \multicolumn{1}{l}{} &  \\
Korean & 39.39 & 16.67 & 14.81 & 16.67 \\
Latvian & 30.30 & 6.67 & 4.00 &  \\
Lithuanian & 31.31 & 6.67 & 14.81 &  \\
Macedonian & 31.31 & 0.00 & 7.41 &  \\
Malayalam & 27.27 & 13.33 & \multicolumn{1}{l}{} &  \\
Marathi & 33.33 & 13.33 & \multicolumn{1}{l}{} &  \\
Nepali & 35.35 & 13.33 & \multicolumn{1}{l}{} &  \\
Norwegian & 37.37 & 16.67 & 11.11 &  \\
Persian & 29.29 & 20.00 & \multicolumn{1}{l}{} &  \\
Polish & 38.38 & 6.67 & 11.11 & 13.33 \\
Portuguese & 47.47 & 20.00 & 8.00 &  \\
Punjabi & 29.29 & 16.67 & \multicolumn{1}{l}{} &  \\
Romanian & 41.41 & 10.00 & 18.52 &  \\
Russian & 46.46 & 16.67 & 12.00 & 20.00 \\
Slovak & 35.35 & 16.67 & 11.11 & 10.00 \\
Slovenian & 35.35 & 23.33 & 11.11 &  \\
Somali & 26.26 & 16.67 & \multicolumn{1}{l}{} &  \\
Spanish & 46.46 & 16.67 & 11.11 &  \\
Swahili & 36.36 & 6.67 & \multicolumn{1}{l}{} &  \\
Swedish & 39.39 & 13.33 & 8.00 &  \\
Tagalog & 35.35 & 13.33 & \multicolumn{1}{l}{} &  \\
Tamil & 33.33 & 10.00 & \multicolumn{1}{l}{} &  \\
Telugu & 34.34 & 13.33 & \multicolumn{1}{l}{} &  \\
Thai & 30.30 & 10.00 & 7.41 &  \\
Turkish & 42.42 & 6.67 & 11.11 &  \\
Ukrainian & 35.35 & 3.33 & 11.11 &  \\
Urdu & 28.28 & 13.33 & \multicolumn{1}{l}{} &  \\
Vietnamese & 31.31 & 10.00 & 7.41 &  \\
Welsh & 30.30 & 23.33 & \multicolumn{1}{l}{} &  \\
\rowcolor[HTML]{FCE5CD} 
English & 65.66 & 20.00 & 25.93 & 53.33 \\ \midrule
Average & 37.47 & 14.85 & 10.50 & 18.40 \\
Standard Deviation & 7.56 & 6.69 & 5.16 & 14.92 \\
Fleiss' Kappa & 0.41 & 0.13 & 0.19 & \\ \bottomrule
\end{tabular}
\caption{\footnotesize Evaluation results of Qwen2.5-Math-1.5B-Instruct + SFT on MCLM.}
\label{tab:qwen_math_1_5B_sft}
\end{table*}

\begin{table*}[]
\centering
\fontsize{9}{11} \selectfont
\begin{tabular}{c|cccc}
\toprule
\textbf{Language} & \textbf{MT-MATH100} & \textbf{MT-AIME2024} & \textbf{M-IMO} & \textbf{M-MO} \\ \midrule
Afrikaans & 39.39 & 10.00 & 13.64 &  \\
Albanian & 39.39 & 16.67 & 7.69 &  \\
Arabic & 41.41 & 16.67 & 14.81 &  \\
Bengali & 39.39 & 30.00 & \multicolumn{1}{l}{} &  \\
Bulgarian & 42.42 & 10.00 & 11.11 &  \\
Catalan & 51.52 & 26.67 & \multicolumn{1}{l}{} &  \\
Chinese (Simplified) & 50.51 & 23.33 & 7.41 &  \\
Chinese (Traditional) & 52.53 & 20.00 & 11.11 & \multirow{-2}{*}{13.33} \\
Croatian & 38.38 & 13.33 & 11.11 &  \\
Czech & 51.52 & 23.33 & 11.11 & 10.00 \\
Danish & 40.40 & 6.67 & 3.70 &  \\
Dutch & 48.48 & 20.00 & 11.11 & 20.00 \\
Estonian & 37.37 & 23.33 & 15.38 &  \\
Finnish & 40.40 & 20.00 & 7.69 &  \\
French & 46.46 & 10.00 & 7.41 & 32.26 \\
German & 49.49 & 10.00 & 7.41 & 3.33 \\
Greek & 28.28 & 20.00 & 17.39 &  \\
Gujarati & 42.42 & 13.33 & \multicolumn{1}{l}{} &  \\
Hebrew & 39.39 & 13.33 & 3.70 &  \\
Hindi & 45.45 & 13.33 & \multicolumn{1}{l}{} &  \\
Hungarian & 43.43 & 40.00 & 11.54 &  \\
Indonesian & 51.52 & 16.67 & 16.67 &  \\
Italian & 48.48 & 13.33 & 11.54 &  \\
Japanese & 50.51 & 6.67 & 11.54 & 3.57 \\
Kannada & 32.32 & 10.00 & \multicolumn{1}{l}{} &  \\
Korean & 55.56 & 10.00 & 11.11 & 26.67 \\
Latvian & 42.42 & 10.00 & 15.38 &  \\
Lithuanian & 36.36 & 13.33 & 7.41 &  \\
Macedonian & 39.39 & 13.33 & 18.52 &  \\
Malayalam & 34.34 & 26.67 & \multicolumn{1}{l}{} &  \\
Marathi & 37.37 & 23.33 & \multicolumn{1}{l}{} &  \\
Nepali & 42.42 & 16.67 & \multicolumn{1}{l}{} &  \\
Norwegian & 42.42 & 10.00 & 3.70 &  \\
Persian & 47.47 & 10.00 & \multicolumn{1}{l}{} &  \\
Polish & 38.38 & 10.00 & 14.81 & 20.00 \\
Portuguese & 50.51 & 26.67 & 11.54 &  \\
Punjabi & 29.29 & 16.67 & \multicolumn{1}{l}{} &  \\
Romanian & 45.45 & 6.67 & 11.11 &  \\
Russian & 57.58 & 13.33 & 7.69 & 36.67 \\
Slovak & 47.47 & 20.00 & 7.41 &  \\
Slovenian & 39.39 & 23.33 & 18.52 &  \\
Somali & 22.22 & 26.67 & \multicolumn{1}{l}{} &  \\
Spanish & 44.44 & 16.67 & 0.00 &  \\
Swahili & 34.34 & 6.67 & \multicolumn{1}{l}{} &  \\
Swedish & 42.42 & 10.00 & 3.85 &  \\
Tagalog & 35.35 & 6.67 & \multicolumn{1}{l}{} &  \\
Tamil & 36.36 & 23.33 & \multicolumn{1}{l}{} &  \\
Telugu & 36.36 & 13.33 & \multicolumn{1}{l}{} &  \\
Thai & 34.34 & 26.67 & 14.81 &  \\
Turkish & 39.39 & 23.33 & 7.41 &  \\
Ukrainian & 49.49 & 10.00 & 7.41 &  \\
Urdu & 32.32 & 20.00 & \multicolumn{1}{l}{} &  \\
Vietnamese & 47.47 & 10.00 & 18.52 &  \\
Welsh & 28.28 & 20.00 & \multicolumn{1}{l}{} &  \\
\rowcolor[HTML]{FCE5CD} 
English & 51.52 & 26.67 & 7.41 & 40.00 \\ \midrule
Average & 42.02 & 16.67 & 10.52 & 20.58 \\
Standard Deviation & 7.46 & 7.31 & 4.63 & 13.17 \\
Fleiss' Kappa & 0.40 & 0.13 & 0.25 & \\ \bottomrule
\end{tabular}
\caption{\footnotesize Evaluation results of Qwen2.5-Math-1.5B-Instruct + MT-SFT on MCLM.}
\label{tab:qwen_math_1_5B_mt_sft}
\end{table*}

\begin{table*}[]
\centering
\fontsize{9}{11} \selectfont
\begin{tabular}{c|cccc}
\toprule
\textbf{Language} & \textbf{MT-MATH100} & \textbf{MT-AIME2024} & \textbf{M-IMO} & \textbf{M-MO} \\ \midrule
Afrikaans & 58.59 & 20.00 & 11.11 &  \\
Albanian & 46.46 & 30.00 & 16.00 &  \\
Arabic & 51.52 & 20.00 & 18.52 &  \\
Bengali & 56.57 & 10.00 & \multicolumn{1}{l}{} &  \\
Bulgarian & 57.58 & 16.67 & 11.11 &  \\
Catalan & 64.65 & 30.00 & \multicolumn{1}{l}{} &  \\
Chinese (Simplified) & 69.70 & 16.67 & 25.93 &  \\
Chinese (Traditional) & 67.68 & 20.00 & 18.52 & \multirow{-2}{*}{33.33} \\
Croatian & 59.60 & 36.67 & 18.52 &  \\
Czech & 57.58 & 33.33 & 18.52 & 16.67 \\
Danish & 56.57 & 16.67 & 14.81 &  \\
Dutch & 64.65 & 30.00 & 22.22 & 23.33 \\
Estonian & 39.39 & 6.67 & 12.00 &  \\
Finnish & 52.53 & 16.67 & 20.00 &  \\
French & 63.64 & 26.67 & 29.63 & 48.39 \\
German & 63.64 & 16.67 & 25.93 & 26.67 \\
Greek & 38.38 & 13.33 & 10.53 &  \\
Gujarati & 47.47 & 3.33 & \multicolumn{1}{l}{} &  \\
Hebrew & 61.62 & 23.33 & 7.41 &  \\
Hindi & 61.62 & 23.33 & \multicolumn{1}{l}{} &  \\
Hungarian & 55.56 & 26.67 & 24.00 &  \\
Indonesian & 69.70 & 13.33 & 23.81 &  \\
Italian & 69.70 & 36.67 & 28.00 &  \\
Japanese & 62.63 & 16.67 & 12.00 & 3.57 \\
Kannada & 42.42 & 16.67 & \multicolumn{1}{l}{} &  \\
Korean & 61.62 & 20.00 & 11.11 & 30.00 \\
Latvian & 49.49 & 6.67 & 20.00 &  \\
Lithuanian & 40.40 & 23.33 & 14.81 &  \\
Macedonian & 59.60 & 23.33 & 25.93 &  \\
Malayalam & 41.41 & 3.33 & \multicolumn{1}{l}{} &  \\
Marathi & 39.39 & 23.33 & \multicolumn{1}{l}{} &  \\
Nepali & 50.51 & 10.00 & \multicolumn{1}{l}{} &  \\
Norwegian & 67.68 & 13.33 & 18.52 &  \\
Persian & 61.62 & 13.33 & \multicolumn{1}{l}{} &  \\
Polish & 62.63 & 16.67 & 22.22 & 23.33 \\
Portuguese & 75.76 & 23.33 & 16.00 &  \\
Punjabi & 42.42 & 13.33 & \multicolumn{1}{l}{} &  \\
Romanian & 58.59 & 26.67 & 22.22 &  \\
Russian & 68.69 & 33.33 & 20.00 & 26.67 \\
Slovak & 58.59 & 13.33 & 11.11 & 20.00 \\
Slovenian & 56.57 & 30.00 & 14.81 &  \\
Somali & 30.30 & 20.00 & \multicolumn{1}{l}{} &  \\
Spanish & 69.70 & 30.00 & 25.93 &  \\
Swahili & 42.42 & 20.00 & \multicolumn{1}{l}{} &  \\
Swedish & 54.55 & 13.33 & 20.00 &  \\
Tagalog & 47.47 & 23.33 & \multicolumn{1}{l}{} &  \\
Tamil & 40.40 & 16.67 & \multicolumn{1}{l}{} &  \\
Telugu & 36.36 & 23.33 & \multicolumn{1}{l}{} &  \\
Thai & 59.60 & 13.33 & 29.63 &  \\
Turkish & 61.62 & 36.67 & 22.22 &  \\
Ukrainian & 67.68 & 16.67 & 18.52 &  \\
Urdu & 50.51 & 20.00 & \multicolumn{1}{l}{} &  \\
Vietnamese & 61.62 & 13.33 & 33.33 &  \\
Welsh & 34.34 & 16.67 & \multicolumn{1}{l}{} &  \\
\rowcolor[HTML]{FCE5CD} 
English & 67.68 & 20.00 & 14.81 & 66.67 \\ \midrule
Average & 55.61 & 19.94 & 19.20 & 28.97 \\
Standard Deviation & 10.93 & 8.10 & 6.24 & 16.64 \\
Fleiss' Kappa & 0.47 & 0.30 & 0.19 & \\ \bottomrule
\end{tabular}
\caption{\footnotesize Evaluation results of DeepSeek-R1-1.5B + MT-SFT on MCLM.}
\label{tab:deepseek_r1_1_5B_mt_sft}
\end{table*}

\begin{table*}[]
\centering
\fontsize{7}{10} \selectfont
\begin{tabular}{c|c|c|cccc}
\toprule
 & \textbf{BF (N=2048)} & \textbf{BF (N=4096)} & \multicolumn{4}{c}{\textbf{BF (N=8192)}} \\ \cmidrule{2-7}
\multirow{-2}{*}{\textbf{Language}} & \multicolumn{1}{c|}{\textbf{MT-AIME2024}} & \multicolumn{1}{c|}{\textbf{MT-AIME2024}} & \textbf{MT-MATH100} & \textbf{MT-AIME2024} & \textbf{M-IMO} & \textbf{M-MO} \\ \midrule
Afrikaans & 23.33 & 23.33 & 59.60 & 30.00 & 9.09 &  \\
Albanian & 23.33 & 26.67 & 48.48 & 26.67 & 7.69 &  \\
Arabic & 16.67 & 23.33 & 60.61 & 26.67 & 14.81 &  \\
Bengali & 33.33 & 30.00 & 54.55 & 23.33 & \multicolumn{1}{l}{} &  \\
Bulgarian & 33.33 & 33.33 & 61.62 & 26.67 & 22.22 &  \\
Catalan & 20.00 & 43.33 & 64.65 & 43.33 & \multicolumn{1}{l}{} &  \\
Chinese (Simplified) & 20.00 & 16.67 & 69.70 & 16.67 & 22.22 &  \\
Chinese (Traditional) & 26.67 & 26.67 & 70.71 & 36.67 & 18.52 & \multirow{-2}{*}{40.00} \\
Croatian & 30.00 & 30.00 & 60.61 & 30.00 & 37.04 &  \\
Czech & 40.00 & 20.00 & 62.63 & 20.00 & 29.63 & 33.33 \\
Danish & 30.00 & 33.33 & 61.62 & 30.00 & 22.22 &  \\
Dutch & 10.00 & 23.33 & 70.71 & 36.67 & 25.93 & 20.00 \\
Estonian & 23.33 & 16.67 & 40.40 & 20.00 & 15.38 &  \\
Finnish & 20.00 & 33.33 & 51.52 & 20.00 & 30.77 &  \\
French & 16.67 & 23.33 & 72.73 & 16.67 & 25.93 & 51.61 \\
German & 26.67 & 20.00 & 75.76 & 26.67 & 25.93 & 30.00 \\
Greek & 6.67 & 13.33 & 42.42 & 16.67 & 21.74 &  \\
Gujarati & 16.67 & 16.67 & 51.52 & 16.67 & \multicolumn{1}{l}{} &  \\
Hebrew & 33.33 & 23.33 & 60.61 & 16.67 & 14.81 &  \\
Hindi & 26.67 & 10.00 & 61.62 & 20.00 & \multicolumn{1}{l}{} &  \\
Hungarian & 30.00 & 26.67 & 58.59 & 23.33 & 26.92 &  \\
Indonesian & 10.00 & 30.00 & 73.74 & 30.00 & 25 &  \\
Italian & 20.00 & 26.67 & 74.75 & 36.67 & 23.08 &  \\
Japanese & 20.00 & 16.67 & 63.64 & 36.67 & 23.08 & 7.14 \\
Kannada & 10.00 & 13.33 & 49.49 & 10.00 & \multicolumn{1}{l}{} &  \\
Korean & 16.67 & 23.33 & 64.65 & 20.00 & 11.11 & 40.00 \\
Latvian & 30.00 & 20.00 & 52.53 & 10.00 & 23.08 &  \\
Lithuanian & 10.00 & 6.67 & 46.46 & 26.67 & 18.52 &  \\
Macedonian & 20.00 & 20.00 & 63.64 & 23.33 & 25.93 &  \\
Malayalam & 10.00 & 13.33 & 51.52 & 13.33 & \multicolumn{1}{l}{} &  \\
Marathi & 20.00 & 26.67 & 51.52 & 23.33 & \multicolumn{1}{l}{} &  \\
Nepali & 30.00 & 13.33 & 54.55 & 20.00 & \multicolumn{1}{l}{} &  \\
Norwegian & 26.67 & 26.67 & 65.66 & 20.00 & 18.52 &  \\
Persian & 26.67 & 23.33 & 62.63 & 36.67 & \multicolumn{1}{l}{} &  \\
Polish & 23.33 & 20.00 & 66.67 & 16.67 & 14.81 & 23.33 \\
Portuguese & 20.00 & 26.67 & 79.80 & 20.00 & 15.38 &  \\
Punjabi & 23.33 & 26.67 & 51.52 & 20.00 & \multicolumn{1}{l}{} &  \\
Romanian & 30.00 & 23.33 & 60.61 & 10.00 & 22.22 &  \\
Russian & 36.67 & 30.00 & 72.73 & 30.00 & 23.08 & 30.00 \\
Slovak & 40.00 & 23.33 & 66.67 & 30.00 & 25.93 &  \\
Slovenian & 20.00 & 20.00 & 60.61 & 33.33 & 25.93 &  \\
Somali & 20.00 & 16.67 & 35.35 & 16.67 & \multicolumn{1}{l}{} &  \\
Spanish & 30.00 & 30.00 & 71.72 & 40.00 & 18.52 &  \\
Swahili & 13.33 & 13.33 & 41.41 & 30.00 & \multicolumn{1}{l}{} &  \\
Swedish & 13.33 & 16.67 & 62.63 & 23.33 & 19.23 &  \\
Tagalog & 10.00 & 20.00 & 52.53 & 23.33 & \multicolumn{1}{l}{} &  \\
Tamil & 26.67 & 20.00 & 44.44 & 23.33 & \multicolumn{1}{l}{} &  \\
Telugu & 13.33 & 16.67 & 44.44 & 20.00 & \multicolumn{1}{l}{} &  \\
Thai & 26.67 & 13.33 & 64.65 & 23.33 & 11.11 &  \\
Turkish & 20.00 & 16.67 & 61.62 & 16.67 & 33.33 &  \\
Ukrainian & 30.00 & 26.67 & 73.74 & 23.33 & 22.22 &  \\
Urdu & 23.33 & 20.00 & 46.46 & 20.00 & \multicolumn{1}{l}{} &  \\
Vietnamese & 20.00 & 26.67 & 62.63 & 40.00 & 25.93 &  \\
Welsh & 20.00 & 16.67 & 42.42 & 13.33 & \multicolumn{1}{l}{} &  \\
\rowcolor[HTML]{FCE5CD}English & 20.00 & 26.67 & 71.72 & 40.00 & 22.22 & 76.67 \\ \midrule
Average & 22.48 & 22.24 & 59.45 & 24.42 & 21.55 & 35.21 \\
Standard Deviation & 7.94 & 6.85 & 10.52 & 8.32 & 6.44 & 19.01 \\
Fleiss' Kappa & 0.33 & 0.37 & 0.44 & 0.32 & 0.19 & \\ \bottomrule
\end{tabular}
\caption{\footnotesize Evaluation results of Qwen2.5-Math-1.5B-Instruct with Budget Forcing (\(BF=2048, 4096, 8192\)).}
\label{tab:1_5B_bf}
\end{table*}

\begin{table*}[]
\centering
\fontsize{9}{11} \selectfont
\begin{tabular}{c|cccc}
\toprule
\textbf{Language} & \textbf{MT-MATH100} & \textbf{MT-AIME2024} & \textbf{M-IMO} & \textbf{M-MO} \\ \midrule
Afrikaans & 72.73 & 13.33 & 27.78 &  \\
Albanian & 60.61 & 16.67 & 20 &  \\
Arabic & 76.77 & 13.33 & 14.81 &  \\
Bengali & 72.73 & 16.67 & \multicolumn{1}{l}{} &  \\
Bulgarian & 72.73 & 16.67 & \multicolumn{1}{l}{} &  \\
Catalan & 73.74 & 20.00 & \multicolumn{1}{l}{} &  \\
Chinese (Simplified) & 77.78 & 20.00 & 7.41 &  \\
Chinese (Traditional) & 73.74 & 23.33 & 11.11 & \multirow{-2}{*}{56.67} \\
Croatian & 73.74 & 30.00 & 22.22 &  \\
Czech & 75.76 & 20.00 & 11.11 & 16.67 \\
Danish & 72.73 & 23.33 & 18.52 &  \\
Dutch & 77.78 & 16.67 & 18.52 & 23.33 \\
Estonian & 57.58 & 13.33 & 20 &  \\
Finnish & 70.71 & 20.00 & 16 &  \\
French & 77.78 & 20.00 & 25.93 & 48.39 \\
German & 76.77 & 23.33 & 25.93 & 26.67 \\
Greek & 64.65 & 13.33 & 10.53 &  \\
Gujarati & 55.56 & 16.67 & \multicolumn{1}{l}{} &  \\
Hebrew & 71.72 & 20.00 & 7.41 &  \\
Hindi & 70.71 & 30.00 & \multicolumn{1}{l}{} &  \\
Hungarian & 71.72 & 26.67 & 20 &  \\
Indonesian & 69.70 & 20.00 & 19.05 &  \\
Italian & 78.79 & 23.33 & 12 &  \\
Japanese & 76.77 & 23.33 & 16 & 3.57 \\
Kannada & 57.58 & 20.00 & \multicolumn{1}{l}{} & 40 \\
Korean & 77.78 & 20.00 & 14.81 &  \\
Latvian & 59.60 & 13.33 & 20 &  \\
Lithuanian & 61.62 & 16.67 & 25.93 &  \\
Macedonian & 77.78 & 16.67 & 22.22 &  \\
Malayalam & 56.57 & 10.00 & \multicolumn{1}{l}{} &  \\
Marathi & 63.64 & 16.67 & \multicolumn{1}{l}{} &  \\
Nepali & 67.68 & 20.00 & \multicolumn{1}{l}{} &  \\
Norwegian & 73.74 & 23.33 & 22.22 &  \\
Persian & 74.75 & 30.00 & \multicolumn{1}{l}{} &  \\
Polish & 71.72 & 16.67 & 22.22 & 26.67 \\
Portuguese & 78.79 & 26.67 & 20 &  \\
Punjabi & 58.59 & 16.67 & \multicolumn{1}{l}{} &  \\
Romanian & 76.77 & 23.33 & 14.81 &  \\
Russian & 77.78 & 20.00 & 20 & 43.33 \\
Slovak & 74.75 & 23.33 & 18.52 & 23.33 \\
Slovenian & 71.72 & 23.33 & 14.81 &  \\
Somali & 38.38 & 6.67 & \multicolumn{1}{l}{} &  \\
Spanish & 75.76 & 30.00 & 14.81 &  \\
Swahili & 46.46 & 13.33 & \multicolumn{1}{l}{} &  \\
Swedish & 76.77 & 16.67 & 24 &  \\
Tagalog & 60.61 & 16.67 & \multicolumn{1}{l}{} &  \\
Tamil & 54.55 & 10.00 & \multicolumn{1}{l}{} &  \\
Telugu & 60.61 & 16.67 & \multicolumn{1}{l}{} &  \\
Thai & 73.74 & 20.00 & 14.81 &  \\
Turkish & 70.71 & 20.00 & 7.41 &  \\
Ukrainian & 76.77 & 23.33 & 14.81 &  \\
Urdu & 63.64 & 50.00 & \multicolumn{1}{l}{} &  \\
Vietnamese & 76.77 & 26.67 & 14.81 &  \\
Welsh & 50.51 & 20.00 & \multicolumn{1}{l}{} &  \\
\rowcolor[HTML]{FCE5CD} 
English & 83.84 & 20.00 & 22.22 & 46.67 \\ \midrule
Average & 69.33 & 20.12 & 17.64 & 32.30 \\
Standard Deviation & 9.42 & 6.57 & 5.38 & 15.92 \\
Fleiss Kappa & 0.61 & 0.51 & 0.38 & 15.81 \\ \bottomrule
\end{tabular}
\caption{\footnotesize Evaluation results of Qwen2.5-Math-7B-Instruct with greedy decoding on MCLM.}
\label{tab:7B_greedy}
\end{table*}

\begin{table*}[]
\centering
\fontsize{7}{10} \selectfont
\begin{tabular}{c|cc|cc|cc}
\toprule
 & \multicolumn{2}{c|}{\textbf{ORM (K=2)}} & \multicolumn{2}{c|}{\textbf{ORM (K=4)}} & \multicolumn{2}{c}{\textbf{ORM (K=8)}} \\ \cmidrule{2-7}
\multirow{-2}{*}{\textbf{Language}} & \textbf{MT-MATH100} & \textbf{MT-AIME2024} & \textbf{MT-MATH100} & \textbf{MT-AIME2024} & \textbf{MT-MATH100} & \textbf{MT-AIME2024} \\ \midrule
Afrikaans & 74.75 & 16.67 & 73.74 & 26.67 & 76.77 & 33.33 \\
Albanian & 68.69 & 20.00 & 65.66 & 26.67 & 68.69 & 26.67 \\
Arabic & 76.77 & 13.33 & 82.83 & 23.33 & 83.84 & 20.00 \\
Bengali & 69.70 & 16.67 & 75.76 & 16.67 & 74.75 & 16.67 \\
Bulgarian & 73.74 & 16.67 & 77.78 & 20.00 & 79.80 & 16.67 \\
Catalan & 75.76 & 26.67 & 77.78 & 20.00 & 76.77 & 30.00 \\
Chinese\_(Simplified) & 77.78 & 20.00 & 81.82 & 26.67 & 82.83 & 26.67 \\
Chinese\_(Traditional) & 77.78 & 23.33 & 81.82 & 23.33 & 81.82 & 23.33 \\
Croatian & 75.76 & 30.00 & 78.79 & 33.33 & 78.79 & 33.33 \\
Czech & 75.76 & 20.00 & 81.82 & 23.33 & 81.82 & 23.33 \\
Danish & 73.74 & 26.67 & 72.73 & 43.33 & 74.75 & 43.33 \\
Dutch & 76.77 & 20.00 & 78.79 & 26.67 & 81.82 & 40.00 \\
Estonian & 62.63 & 16.67 & 64.65 & 23.33 & 65.66 & 30.00 \\
Finnish & 73.74 & 23.33 & 77.78 & 33.33 & 75.76 & 33.33 \\
French & 81.82 & 23.33 & 81.82 & 20.00 & 81.82 & 26.67 \\
German & 78.79 & 33.33 & 81.82 & 40.00 & 83.84 & 40.00 \\
Greek & 65.66 & 20.00 & 67.68 & 23.33 & 70.71 & 16.67 \\
Gujarati & 58.59 & 13.33 & 59.60 & 20.00 & 64.65 & 16.67 \\
Hebrew & 73.74 & 13.33 & 75.76 & 20.00 & 76.77 & 30.00 \\
Hindi & 70.71 & 26.67 & 75.76 & 26.67 & 75.76 & 36.67 \\
Hungarian & 73.74 & 26.67 & 76.77 & 20.00 & 76.77 & 23.33 \\
Indonesian & 75.76 & 30.00 & 76.77 & 33.33 & 77.78 & 43.33 \\
Italian & 79.80 & 26.67 & 79.80 & 26.67 & 82.83 & 33.33 \\
Japanese & 78.79 & 23.33 & 79.80 & 30.00 & 80.81 & 23.33 \\
Kannada & 55.56 & 13.33 & 57.58 & 13.33 & 59.60 & 20.00 \\
Korean & 79.80 & 16.67 & 76.77 & 23.33 & 77.78 & 26.67 \\
Latvian & 61.62 & 16.67 & 65.66 & 10.00 & 66.67 & 10.00 \\
Lithuanian & 63.64 & 20.00 & 68.69 & 30.00 & 69.70 & 20.00 \\
Macedonian & 76.77 & 16.67 & 80.81 & 20.00 & 79.80 & 23.33 \\
Malayalam & 59.60 & 10.00 & 62.63 & 16.67 & 68.69 & 23.33 \\
Marathi & 65.66 & 26.67 & 68.69 & 20.00 & 69.70 & 16.67 \\
Nepali & 64.65 & 13.33 & 69.70 & 16.67 & 68.69 & 16.67 \\
Norwegian & 72.73 & 26.67 & 74.75 & 30.00 & 76.77 & 33.33 \\
Persian & 76.77 & 23.33 & 75.76 & 23.33 & 76.77 & 16.67 \\
Polish & 77.78 & 10.00 & 78.79 & 10.00 & 78.79 & 16.67 \\
Portuguese & 81.82 & 26.67 & 80.81 & 36.67 & 83.84 & 40.00 \\
Punjabi & 58.59 & 20.00 & 59.60 & 16.67 & 62.63 & 26.67 \\
Romanian & 79.80 & 23.33 & 81.82 & 26.67 & 79.80 & 30.00 \\
Russian & 78.79 & 26.67 & 82.83 & 20.00 & 86.87 & 26.67 \\
Slovak & 77.78 & 30.00 & 79.80 & 33.33 & 81.82 & 30.00 \\
Slovenian & 73.74 & 13.33 & 78.79 & 20.00 & 78.79 & 23.33 \\
Somali & 38.38 & 6.67 & 42.42 & 13.33 & 44.44 & 20.00 \\
Spanish & 75.76 & 26.67 & 78.79 & 26.67 & 81.82 & 30.00 \\
Swahili & 48.48 & 13.33 & 49.49 & 20.00 & 51.52 & 23.33 \\
Swedish & 77.78 & 30.00 & 76.77 & 30.00 & 77.78 & 30.00 \\
Tagalog & 58.59 & 13.33 & 65.66 & 10.00 & 66.67 & 16.67 \\
Tamil & 59.60 & 16.67 & 65.66 & 10.00 & 62.63 & 10.00 \\
Telugu & 61.62 & 20.00 & 63.64 & 23.33 & 62.63 & 16.67 \\
Thai & 76.77 & 16.67 & 79.80 & 23.33 & 77.78 & 30.00 \\
Turkish & 76.77 & 26.67 & 79.80 & 26.67 & 79.80 & 26.67 \\
Ukrainian & 77.78 & 23.33 & 78.79 & 23.33 & 79.80 & 26.67 \\
Urdu & 66.67 & 33.33 & 67.68 & 30.00 & 72.73 & 30.00 \\
Vietnamese & 73.74 & 33.33 & 76.77 & 33.33 & 80.81 & 36.67 \\
Welsh & 51.52 & 20.00 & 53.54 & 16.67 & 56.57 & 6.67 \\
\rowcolor[HTML]{FCE5CD} 
English & 84.85 & 26.67 & 84.85 & 30.00 & 86.87 & 26.67 \\ \midrule
Average & 70.98 & 21.21 & 73.35 & 23.82 & 74.62 & 25.76 \\
Standard Deviation & 9.46 & 6.52 & 9.20 & 7.41 & 8.86 & 8.37 \\
Fleiss' Kappa & 0.62 & 0.55 & 0.65 & 0.57 & 0.67 & 0.57 \\ \bottomrule
\end{tabular}
\caption{\footnotesize Evaluation results of Qwen2.5-Math-7B-Instruct with Best-of-N \((K=2, 4, 8)\) using Qwen2.5-Math-RM-72B as ORM on MT-MATH100 and MT-AIME2024.}
\label{tab:7B_orm_72B}
\end{table*}

\begin{table*}[]
\centering
\fontsize{7}{10} \selectfont
\begin{tabular}{c|cc|cc|cc}
\toprule
 & \multicolumn{2}{c|}{\textbf{PRM (S=3, c=3)}} & \multicolumn{2}{c|}{\textbf{PRM (S=4, c=5)}} & \multicolumn{2}{c}{\textbf{PRM (S=5, c=8)}} \\ \cmidrule{2-7}
\multirow{-2}{*}{\textbf{Language}} & \textbf{MT-MATH100} & \textbf{MT-AIME2024} & \textbf{MT-MATH100} & \textbf{MT-AIME2024} & \textbf{MT-MATH100} & \textbf{MT-AIME2024} \\ \midrule
Afrikaans & 70.71 & 20.00 & 70.71 & 16.67 & 70.71 & 20.00 \\
Albanian & 60.61 & 16.67 & 62.63 & 33.33 & 61.62 & 26.67 \\
Arabic & 65.66 & 26.67 & 78.79 & 26.67 & 82.83 & 30.00 \\
Bengali & 67.68 & 16.67 & 70.71 & 10.00 & 68.69 & 23.33 \\
Bulgarian & 69.70 & 20.00 & 74.75 & 10.00 & 75.76 & 30.00 \\
Catalan & 72.73 & 16.67 & 70.71 & 20.00 & 71.72 & 16.67 \\
Chinese (Simplified) & 72.73 & 16.67 & 73.74 & 33.33 & 78.79 & 30.00 \\
Chinese (Traditional) & 71.72 & 16.67 & 76.77 & 20.00 & 77.78 & 23.33 \\
Croatian & 69.70 & 20.00 & 72.73 & 16.67 & 70.71 & 33.33 \\
Czech & 69.70 & 16.67 & 77.78 & 10.00 & 73.74 & 30.00 \\
Danish & 63.64 & 23.33 & 69.70 & 33.33 & 66.67 & 30.00 \\
Dutch & 71.72 & 6.67 & 72.73 & 26.67 & 75.76 & 26.67 \\
Estonian & 46.46 & 20.00 & 51.52 & 13.33 & 59.60 & 20.00 \\
Finnish & 64.65 & 16.67 & 66.67 & 13.33 & 72.73 & 33.33 \\
French & 73.74 & 20.00 & 72.73 & 16.67 & 76.77 & 26.67 \\
German & 73.74 & 10.00 & 68.69 & 10.00 & 76.77 & 26.67 \\
Greek & 63.64 & 16.67 & 64.65 & 13.33 & 67.68 & 13.33 \\
Gujarati & 56.57 & 13.33 & 56.57 & 26.67 & 55.56 & 13.33 \\
Hebrew & 66.67 & 10.00 & 68.69 & 20.00 & 75.76 & 26.67 \\
Hindi & 58.59 & 16.67 & 63.64 & 20.00 & 72.73 & 13.33 \\
Hungarian & 68.69 & 16.67 & 69.70 & 30.00 & 72.73 & 20.00 \\
Indonesian & 69.70 & 26.67 & 68.69 & 20.00 & 72.73 & 10.00 \\
Italian & 71.72 & 16.67 & 77.78 & 30.00 & 73.74 & 23.33 \\
Japanese & 71.72 & 23.33 & 75.76 & 16.67 & 76.77 & 13.33 \\
Kannada & 46.46 & 16.67 & 53.54 & 10.00 & 54.55 & 16.67 \\
Korean & 69.70 & 16.67 & 72.73 & 13.33 & 74.75 & 16.67 \\
Latvian & 59.60 & 10.00 & 63.64 & 13.33 & 63.64 & 16.67 \\
Lithuanian & 55.56 & 20.00 & 62.63 & 13.33 & 65.66 & 16.67 \\
Macedonian & 69.70 & 16.67 & 75.76 & 16.67 & 75.76 & 23.33 \\
Malayalam & 49.49 & 20.00 & 57.58 & 23.33 & 52.53 & 20.00 \\
Marathi & 56.57 & 20.00 & 55.56 & 23.33 & 57.58 & 23.33 \\
Nepali & 51.52 & 16.67 & 61.62 & 20.00 & 52.53 & 23.33 \\
Norwegian & 69.70 & 20.00 & 67.68 & 20.00 & 69.70 & 26.67 \\
Persian & 71.72 & 26.67 & 71.72 & 16.67 & 72.73 & 23.33 \\
Polish & 61.62 & 13.33 & 67.68 & 13.33 & 76.77 & 10.00 \\
Portuguese & 72.73 & 10.00 & 71.72 & 26.67 & 79.80 & 26.67 \\
Punjabi & 46.46 & 13.33 & 45.45 & 10.00 & 52.53 & 20.00 \\
Romanian & 66.67 & 13.33 & 70.71 & 33.33 & 77.78 & 30.00 \\
Russian & 75.76 & 16.67 & 76.77 & 16.67 & 76.77 & 33.33 \\
Slovak & 70.71 & 23.33 & 75.76 & 26.67 & 70.71 & 13.33 \\
Slovenian & 70.71 & 23.33 & 72.73 & 30.00 & 74.75 & 20.00 \\
Somali & 40.40 & 3.33 & 42.42 & 10.00 & 42.42 & 6.67 \\
Spanish & 71.72 & 13.33 & 77.78 & 20.00 & 80.81 & 23.33 \\
Swahili & 48.48 & 6.67 & 42.42 & 10.00 & 44.44 & 16.67 \\
Swedish & 70.71 & 16.67 & 76.77 & 36.67 & 71.72 & 26.67 \\
Tagalog & 55.56 & 23.33 & 59.60 & 16.67 & 58.59 & 13.33 \\
Tamil & 50.51 & 10.00 & 55.56 & 3.33 & 57.58 & 20.00 \\
Telugu & 53.54 & 13.33 & 58.59 & 20.00 & 54.55 & 20.00 \\
Thai & 67.68 & 10.00 & 71.72 & 16.67 & 71.72 & 26.67 \\
Turkish & 63.64 & 20.00 & 71.72 & 20.00 & 64.65 & 16.67 \\
Ukrainian & 75.76 & 20.00 & 77.78 & 26.67 & 79.80 & 20.00 \\
Urdu & 57.58 & 26.67 & 62.63 & 20.00 & 66.67 & 23.33 \\
Vietnamese & 72.73 & 23.33 & 73.74 & 13.33 & 73.74 & 33.33 \\
Welsh & 50.51 & 20.00 & 43.43 & 13.33 & 45.45 & 20.00 \\
\rowcolor[HTML]{FCE5CD} 
English & 73.74 & 23.33 & 75.76 & 20.00 & 75.76 & 23.33 \\ \midrule
Average & 64.17 & 17.27 & 67.09 & 19.27 & 68.45 & 22.00 \\
Standard Deviation & 9.25 & 5.33 & 9.65 & 7.61 & 10.02 & 6.56 \\
Fleiss' Kappa & 0.56 & 0.56 & 0.54 & 0.57 & 0.56 & 0.59 \\ \bottomrule
\end{tabular}
\caption{\footnotesize Evaluation results of Qwen2.5-Math-7B-Instruct using Qwen2.5-Math-PRM-72B as PRM on MT-MATH100 and MT-AIME2024.}
\label{tab:7B_prm_72B}
\end{table*}

\begin{table*}[]
\centering
\fontsize{9}{11} \selectfont
\begin{tabular}{c|cccc}
\toprule
\textbf{Language} & \textbf{MT-MATH100} & \textbf{MT-AIME2024} & \textbf{M-IMO} & \textbf{M-MO} \\ \midrule
Afrikaans & 73.74 & 23.33 & 9.09 &  \\
Albanian & 66.67 & 20.00 & 15.38 &  \\
Arabic & 71.72 & 16.67 & 3.70 &  \\
Bengali & 64.65 & 3.33 & \multicolumn{1}{l}{} &  \\
Bulgarian & 72.73 & 20.00 & 18.52 &  \\
Catalan & 70.71 & 26.67 & \multicolumn{1}{l}{} &  \\
Chinese (Simplified) & 70.71 & 23.33 & 14.81 &  \\
Chinese (Traditional) & 69.70 & 23.33 & 11.11 & \multirow{-2}{*}{26.67} \\
Croatian & 72.73 & 16.67 & 18.52 &  \\
Czech & 71.72 & 33.33 & 11.11 & 36.67 \\
Danish & 71.72 & 23.33 & 22.22 &  \\
Dutch & 69.70 & 20.00 & 3.70 & 3.33 \\
Estonian & 76.77 & 16.67 & 15.38 &  \\
Finnish & 72.73 & 6.67 & 15.38 &  \\
French & 70.71 & 23.33 & 14.81 & 48.39 \\
German & 73.74 & 20.00 & 18.52 & 26.67 \\
Greek & 71.72 & 10.00 & 13.04 &  \\
Gujarati & 67.68 & 13.33 & \multicolumn{1}{l}{} &  \\
Hebrew & 71.72 & 10.00 & 7.41 &  \\
Hindi & 70.71 & 6.67 & \multicolumn{1}{l}{} &  \\
Hungarian & 73.74 & 26.67 & 11.54 &  \\
Indonesian & 68.69 & 13.33 & 16.67 &  \\
Italian & 72.73 & 23.33 & 11.54 &  \\
Japanese & 70.71 & 30.00 & 7.69 & 7.14 \\
Kannada & 61.62 & 23.33 & \multicolumn{1}{l}{} &  \\
Korean & 72.73 & 26.67 & 22.22 & 36.67 \\
Latvian & 69.70 & 20.00 & 7.69 &  \\
Lithuanian & 68.69 & 16.67 & 7.41 &  \\
Macedonian & 71.72 & 20.00 & 22.22 &  \\
Malayalam & 62.63 & 23.33 & \multicolumn{1}{l}{} &  \\
Marathi & 63.64 & 20.00 & \multicolumn{1}{l}{} &  \\
Nepali & 67.68 & 10.00 & \multicolumn{1}{l}{} &  \\
Norwegian & 75.76 & 30.00 & 11.11 &  \\
Persian & 66.67 & 26.67 & \multicolumn{1}{l}{} &  \\
Polish & 72.73 & 13.33 & 22.22 & 26.67 \\
Portuguese & 70.71 & 26.67 & 7.69 &  \\
Punjabi & 69.70 & 16.67 & \multicolumn{1}{l}{} &  \\
Romanian & 73.74 & 26.67 & 11.11 &  \\
Russian & 73.74 & 23.33 & 15.38 & 50.00 \\
Slovak & 72.73 & 20.00 & 18.52 &  \\
Slovenian & 72.73 & 16.67 & 7.41 &  \\
Somali & 57.58 & 20.00 & \multicolumn{1}{l}{} &  \\
Spanish & 71.72 & 26.67 & 14.81 &  \\
Swahili & 65.66 & 23.33 & \multicolumn{1}{l}{} &  \\
Swedish & 72.73 & 23.33 & 23.08 &  \\
Tagalog & 71.72 & 20.00 & \multicolumn{1}{l}{} &  \\
Tamil & 67.68 & 20.00 & \multicolumn{1}{l}{} &  \\
Telugu & 66.67 & 16.67 & \multicolumn{1}{l}{} &  \\
Thai & 70.71 & 26.67 & 7.41 &  \\
Turkish & 71.72 & 10.00 & 11.11 &  \\
Ukrainian & 73.74 & 23.33 & 14.81 &  \\
Urdu & 68.69 & 23.33 & \multicolumn{1}{l}{} &  \\
Vietnamese & 71.72 & 6.67 & 14.81 &  \\
Welsh & 65.66 & 26.67 & \multicolumn{1}{l}{} &  \\
\rowcolor[HTML]{FCE5CD} 
English & 75.76 & 33.33 & 7.41 & 50.00 \\ \midrule
Average & 70.30 & 20.18 & 13.33 & 30.81 \\
Standard Deviation & 3.68 & 6.83 & 5.36 & 15.80 \\
Fleiss' Kappa & 0.71 & 0.33 & 0.25 & \\ \bottomrule
\end{tabular}
\caption{\footnotesize Evaluation results of \textsc{gpt-4o-mini} with greedy decoding on MCLM.}
\label{tab:gpt_4o_mini_result}
\end{table*}

\begin{table*}[]
\centering
\fontsize{9}{11} \selectfont
\begin{tabular}{c|cccc}
\toprule
\textbf{Language} & \textbf{MT-MATH100} & \textbf{MT-AIME2024} & \textbf{M-IMO} & \textbf{M-MO} \\ \midrule
Afrikaans & 85.86 & 46.67 & 33.33 &  \\
Albanian & 86.87 & 53.33 & 28.00 &  \\
Arabic & 86.87 & 43.33 & 22.22 &  \\
Bengali & 86.87 & 43.33 & \multicolumn{1}{l}{} &  \\
Bulgarian & 87.88 & 46.67 & 40.74 &  \\
Catalan & 87.88 & 53.33 & \multicolumn{1}{l}{} &  \\
Chinese (Simplified) & 85.86 & 50 & 25.93 &  \\
Chinese (Traditional) & 84.85 & 40 & 29.63 & \multirow{-2}{*}{66.67} \\
Croatian & 84.85 & 46.67 & 33.33 &  \\
Czech & 84.85 & 36.67 & 29.63 & 53.33 \\
Danish & 85.86 & 40 & 40.74 &  \\
Dutch & 86.87 & 50 & 33.33 & 40.00 \\
Estonian & 83.84 & 50 & 28.00 &  \\
Finnish & 84.85 & 40 & 28.00 &  \\
French & 86.87 & 43.33 & 29.63 & 67.74 \\
German & 86.87 & 43.33 & 33.33 & 43.33 \\
Greek & 87.88 & 56.67 & 21.05 &  \\
Gujarati & 83.84 & 46.67 & \multicolumn{1}{l}{} &  \\
Hebrew & 81.82 & 40 & 7.41 &  \\
Hindi & 83.84 & 43.33 & \multicolumn{1}{l}{} &  \\
Hungarian & 86.87 & 53.33 & 28.00 &  \\
Indonesian & 84.85 & 43.33 & 33.33 &  \\
Italian & 82.83 & 50 & 36.00 &  \\
Japanese & 86.87 & 50 & 16.00 & 17.86 \\
Kannada & 86.87 & 43.33 & \multicolumn{1}{l}{} &  \\
Korean & 77.78 & 46.67 & 25.93 & 60.00 \\
Latvian & 87.88 & 46.67 & 32.00 &  \\
Lithuanian & 85.86 & 46.67 & 33.33 &  \\
Macedonian & 83.84 & 43.33 & 33.33 &  \\
Malayalam & 85.86 & 46.67 & \multicolumn{1}{l}{} &  \\
Marathi & 83.84 & 36.67 & \multicolumn{1}{l}{} &  \\
Nepali & 79.8 & 46.67 & \multicolumn{1}{l}{} &  \\
Norwegian & 82.83 & 53.33 & 22.22 &  \\
Persian & 87.88 & 53.33 & \multicolumn{1}{l}{} &  \\
Polish & 81.82 & 43.33 & 37.04 & 40.00 \\
Portuguese & 82.83 & 36.67 & 36.00 &  \\
Punjabi & 87.88 & 43.33 & \multicolumn{1}{l}{} &  \\
Romanian & 81.82 & 40 & 40.74 &  \\
Russian & 85.86 & 56.67 & 20.00 & 50.00 \\
Slovak & 87.88 & 46.67 & 33.33 & 46.67 \\
Slovenian & 85.86 & 46.67 & 29.63 &  \\
Somali & 87.88 & 50 & \multicolumn{1}{l}{} &  \\
Spanish & 72.73 & 50 & 29.63 &  \\
Swahili & 86.87 & 43.33 & \multicolumn{1}{l}{} &  \\
Swedish & 79.8 & 43.33 & 28.00 &  \\
Tagalog & 85.86 & 46.67 & \multicolumn{1}{l}{} &  \\
Tamil & 84.85 & 43.33 & \multicolumn{1}{l}{} &  \\
Telugu & 82.83 & 33.33 & \multicolumn{1}{l}{} &  \\
Thai & 84.85 & 40 & 22.22 &  \\
Turkish & 84.85 & 40 & 33.33 &  \\
Ukrainian & 84.85 & 50 & 29.63 &  \\
Urdu & 84.85 & 36.67 & \multicolumn{1}{l}{} &  \\
Vietnamese & 85.86 & 46.67 & 37.04 &  \\
Welsh & 85.86 & 46.67 & \multicolumn{1}{l}{} &  \\
\rowcolor[HTML]{FCE5CD} 
\cellcolor[HTML]{FCE5CD}English & 83.84 & 36.67 & 29.63 & 80.00 \\ \midrule
Average & 84.89 & 45.33 & 29.75 & 51.42 \\
Standard Deviation & 2.80 & 5.35 & 6.86 & 16.94 \\
Fleiss' Kappa & 0.88 & 0.73 & 0.44 & \\ \bottomrule
\end{tabular}
\caption{\footnotesize Evaluation results of \textsc{o3-mini} with greedy decoding on MCLM.}
\label{tab:o3_mini_result}
\end{table*}

\end{document}